\documentclass[letterpaper]{article} % DO NOT CHANGE THIS
\usepackage{aaai2026}  % DO NOT CHANGE THIS
\usepackage{times}  % DO NOT CHANGE THIS
\usepackage{helvet}  % DO NOT CHANGE THIS
\usepackage{courier}  % DO NOT CHANGE THIS
\usepackage[hyphens]{url}  % DO NOT CHANGE THIS
\usepackage{graphicx} % DO NOT CHANGE THIS
\usepackage{natbib}  % DO NOT CHANGE THIS AND DO NOT ADD ANY OPTIONS TO IT
\usepackage{pifont}
\usepackage{caption} % DO NOT CHANGE THIS AND DO NOT ADD ANY OPTIONS TO IT
\usepackage{multirow}
\usepackage{booktabs}
\usepackage{algorithm}
\usepackage{algorithmic}
\usepackage{makecell}
\usepackage{colortbl}  % 提供\rowcolor命令
\usepackage{xcolor}   
\usepackage{tikz}

\definecolor{lightblue}{RGB}{229,248,255}
\definecolor{gray_venue}{rgb}{0.53,0.52,0.52}
\definecolor{gray}{rgb}{0.9,0.9,0.9}
\usepackage{newfloat}
\usepackage{listings}
\DeclareCaptionStyle{ruled}{labelfont=normalfont,labelsep=colon,strut=off} % DO NOT CHANGE THIS
\floatstyle{ruled}
\newfloat{listing}{tb}{lst}{}
\floatname{listing}{Listing}

\makeatletter
\def\whline#1{%
	\noalign{\ifnum0=`}\fi\hrule \@height #1 \futurelet
	\reserved@a\@xhline}

\usepackage{paralist}

\title{PosterVerse: A Full-Workflow Framework for Commercial-Grade Poster Generation with HTML-Based Scalable Typography}
\author{
    Junle Liu\equalcontrib\textsuperscript{\rm 1,\rm 3}, 
    Peirong Zhang\equalcontrib\textsuperscript{\rm 1}, 
    Yuyi Zhang\textsuperscript{\rm 1,\rm 3}, 
    Pengyu Yan\textsuperscript{\rm 1}, 
    Hui Zhou\textsuperscript{\rm 2,\rm 3}, \\
    Xinyue Zhou\textsuperscript{\rm 2,\rm 3}, 
    Fengjun Guo\textsuperscript{\rm 2,\rm 3}, 
    Lianwen Jin\textsuperscript{\rm 1,\rm 3}\thanks{Corresponding Author.}
}
\affiliations{

    \textsuperscript{\rm 1}South China University of Technology\\
    \textsuperscript{\rm 2}Intsig Information Co., Ltd.\\
    \textsuperscript{\rm 3}INTSIG-SCUT Joint Lab on Document Analysis and Recognition\\
    {junle$\_$liu@foxmail.com}, {eelwjin@scut.edu.cn}
}

\usepackage{bibentry}
\begin{document}

\maketitle

\begin{abstract}
Commercial-grade poster design demands the seamless integration of aesthetic appeal with precise, informative content delivery. Current automated poster generation systems face significant limitations, including incomplete design workflows, poor text rendering accuracy, and insufficient flexibility for commercial applications. To address these challenges, we propose \textbf{PosterVerse}, a full-workflow, commercial-grade poster generation method that seamlessly automates the entire design process while delivering high-density and scalable text rendering. PosterVerse replicates professional design through three key stages: (1) blueprint creation using fine-tuned LLMs to extract key design elements from user requirements, (2) graphical background generation via customized diffusion models to create visually appealing imagery, and (3) unified layout-text rendering with an MLLM-powered HTML engine to guarantee high text accuracy and flexible customization. In addition, we introduce \textbf{PosterDNA}, a commercial-grade, HTML-based dataset tailored for training and validating poster design models. To the best of our knowledge, PosterDNA is the first Chinese poster generation dataset to introduce HTML typography files, enabling scalable text rendering and fundamentally solving the challenges of rendering small and high-density text. Experimental results demonstrate that PosterVerse consistently produces commercial-grade posters with appealing visuals, accurate text alignment, and customizable layouts, making it a promising solution for automating commercial poster design. % Dataset and code will be publicly available.

\end{abstract}

\begin{links}
    \link{Code}{https://github.com/wuhaer/PosterVerse}
\end{links}

\section{Introduction}

Poster design plays a crucial role in business, culture, and marketing. An effective poster must distill complex information into clear, compelling messages while balancing informational richness with focused messaging. This requires sophisticated orchestration of font, color, and layout skills that traditionally demand extensive design expertise and time-intensive manual processes. With the rapid development of AI-generated content (AIGC) technologies~\cite{gemini, flux, gpt4ogen2025}, generative models have become key tools for commercial creation~\cite{autoposter, postermaker}, making their application in automatic design an inevitable trend. 

\begin{figure}[t]
  \includegraphics[width=1\columnwidth]{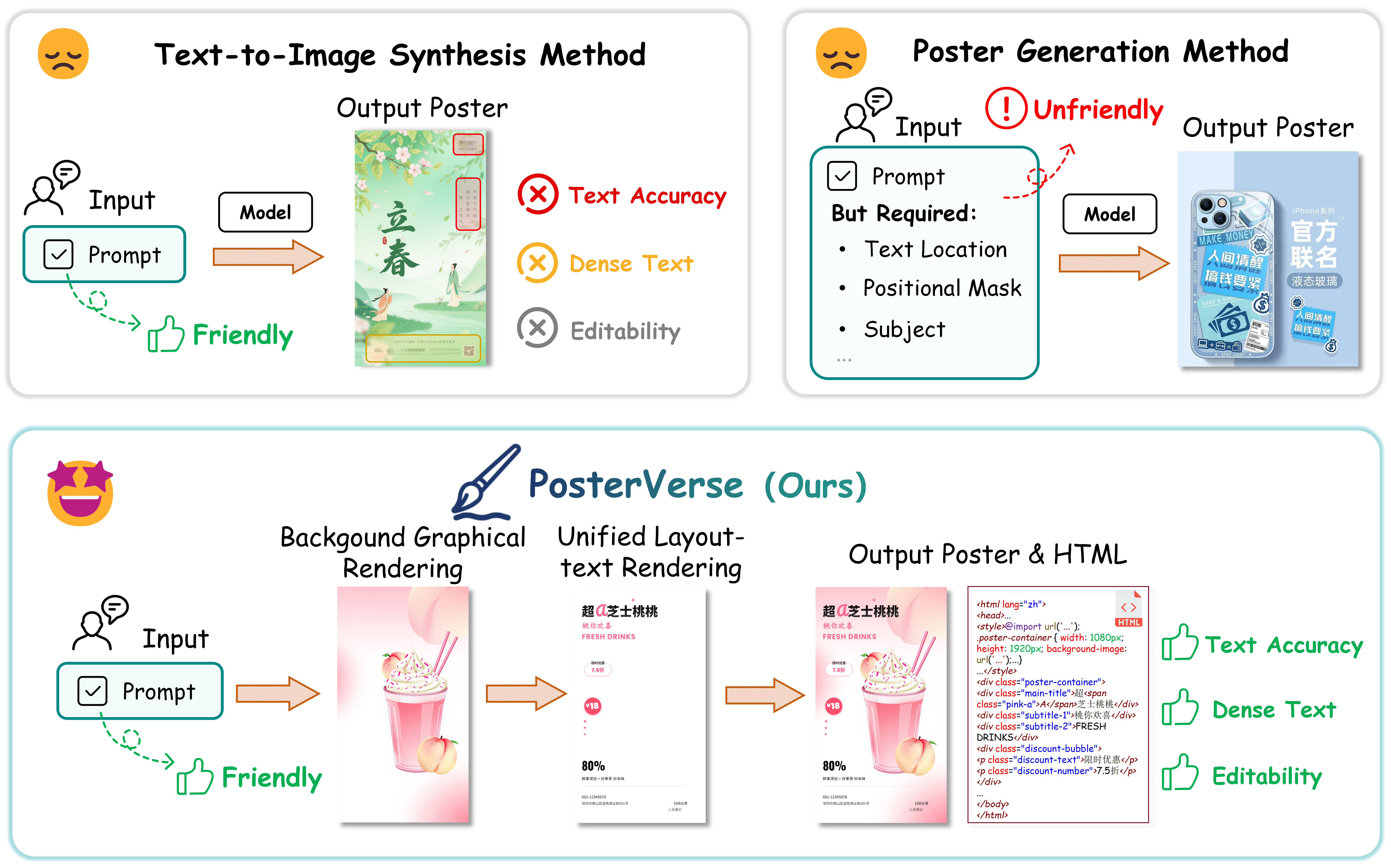}
  \centering
  \caption{Comparison between existing poster generation methods (top) and the proposed PosterVerse (bottom).}
  \vspace{-3mm}
  \label{fig: intro}
\end{figure}

\begin{figure*}[t]
  \includegraphics[width=1\linewidth]{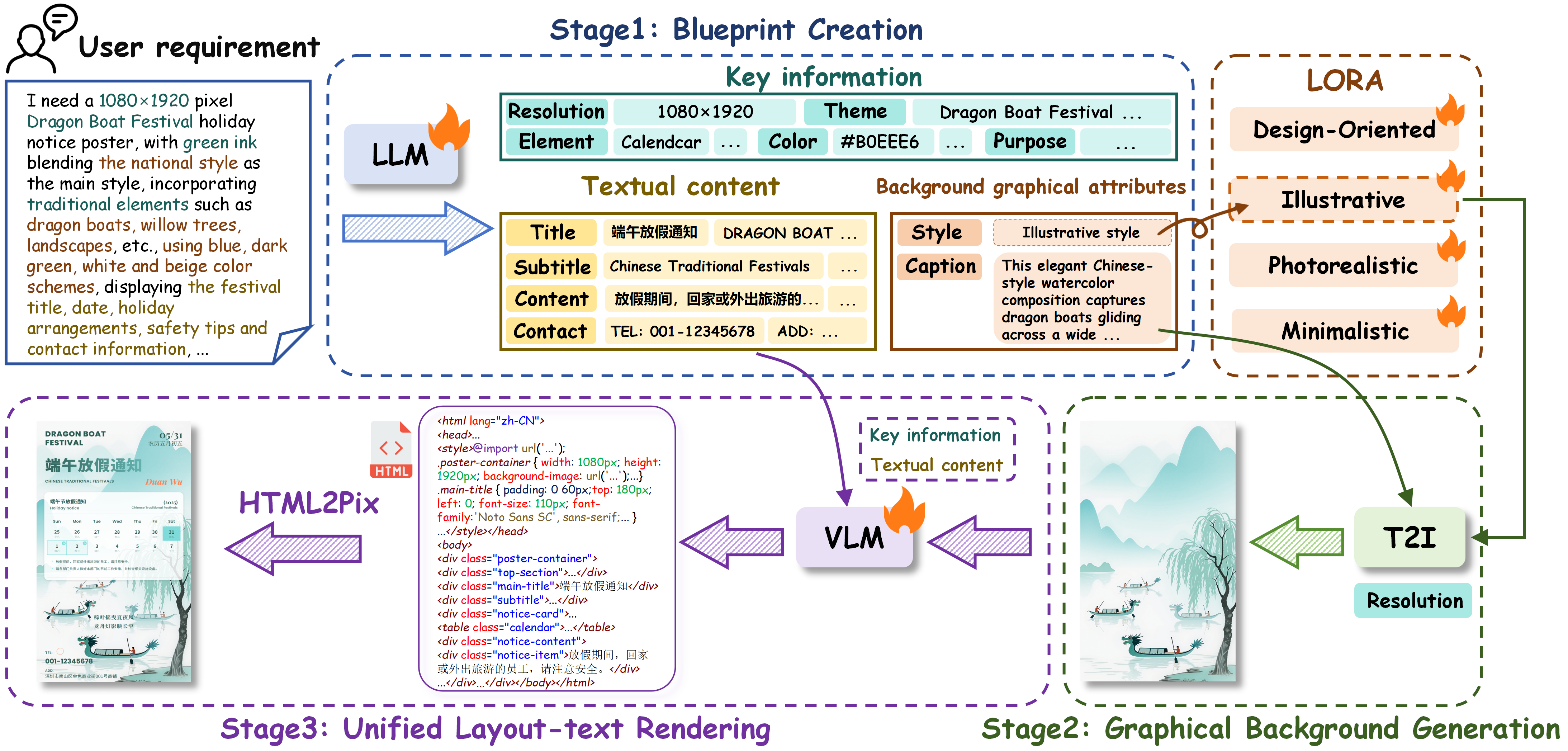}
  \centering
  \caption{The overview of PosterVerse: A full-workflow method integrating blueprint creation, graphical background generation, and unified layout-text rendering to produce commercial-grade, aesthetically appealing, and text-rich posters.}
  \label{fig: posterverse_pipeline}
\end{figure*}

However, existing approaches suffer from several limitations. 
\textbf{(1) Lack of full workflow solutions.} Poster generation involves multiple stages, including background graphic design, layout planning, and font rendering. Yet, many methods~\cite{Posterllama, Relation-Aware-Diffusion-Model, Planning-and-rendering, posterlayout} only focus on partial stages such as generating layout or typography alone, failing to provide full-workflow poster design systems. \textbf{(2) Poor flexibility and accessibility.} While some works~\cite{postermaker, bizgen} demonstrate promising visual effects, they heavily rely on additional inputs beyond textual prompts, such as positional masks, text bounding boxes, and graphical subjects. This greatly hampers their flexibility compared to prompt-only systems, especially for non-technical users. \textbf{(3) Inaccurate text rendering.} Despite the stunning aesthetic creation abilities of models like GPT-4o~\cite{gpt4ogen2025} and Gemini~\cite{gemini}, they typically struggle with generating accurate text, particularly evident in Chinese characters and dense, small-scale characters. The generated texts are usually illegible or semantically meaningless. \textbf{(4) Insufficient understanding of user requirements.} Existing text-to-image models~\cite{flux, sd3.5} are usually constrained by input token limits or primitive understanding capabilities of text encoders such as T5~\cite{t5}, limiting them to fully comprehend user needs. \textbf{(5) Gap to commercial utility.} Current poster design methods~\cite{postercraft, prompt2poster} mainly prioritize artistic expression over user requirement adherence. Although aesthetically appealing, the generated posters are commercially impractical and require substantial manual post-processing. Additionally, most of them generate non-editable static images. This prevents post-adjustments to poster content like text, fonts, or layout, making them ill-suited for the dynamic nature of commercial scenarios where requirements frequently evolve.

To bridge these gaps, we propose \textbf{PosterVerse}, a full-workflow, prompt-driven poster generation framework, featuring scalable text rendering to address small, dense text synthesis and natively editable output for flexible post-editing. Specifically, PosterVerse mimics the process of professional designers creating posters by dividing the poster design process into three stages: 
\textbf{(1) Blueprint creation.} In this stage, we utilize a fine-tuned Large Language Model (LLM) to interpret and expand upon the user's requirement. Irrespective of the initial input's level of detail, this process transforms the request into a comprehensive design specification. We also extract key design elements such as themes, styles, colors, and text content, serving as foundations for the subsequent stages.  \textbf{(2) Graphical background generation.} Building upon Flux.1-dev fine-tuned by LoRA, this stage generates high-quality backgrounds that strictly adhere to the design blueprint from the first stage. To afford users greater creative control, PosterVerse also supports direct upload of custom background images as an alternative. \textbf{(3) Unified layout-text rendering.} In this stage, a multimodal LLM (MLLM) is used to synthesize the final layout, integrating specified text and design elements with the background graphic. The output is a complete HTML document that ensures perfect typographic accuracy, addressing a known weakness in many generative models. Moreover, HTML enables efficient and customizable post-edits, accommodating dynamic design scenarios that require frequent modifications such as commercial.

Furthermore, to address the critical lack of commercial-grade dataset, we present \textbf{PosterDNA}, an HTML-based poster generation dataset with fine-grained specifications. Developed in collaboration with professional designers for high quality and practical relevance, PosterDNA comprises a diverse collection of poster samples characterized by complex layouts and dense textual designs. Each entry is a structured tuple of ``requirements-graphic-layout-poster'', specifically engineered to support the modular training and validation of our PosterVerse. It pioneers in introducing HTML-based typography files, standing as the first Chinese poster design dataset that addresses small, dense text rendering and potentially inspiring future works.

Overall, our contributions can be summarized as follows:
\begin{itemize}
    \item We propose \textbf{PosterDNA}, the first commercial-grade and text-dense poster generation dataset with fine-grained HTML-based specifications, designed to support modular training and validation with high-quality samples.
    \item We propose \textbf{PosterVerse}, a full-workflow method that integrates blueprint creation, graphical background generation, and unified layout-text rendering, enabling the creation of posters with aesthetically sophisticated layouts and text-dense designs for commercial-grade use.
    \item PosterVerse allows users to generate commercial-grade posters solely from textual prompts.
    % , while maintaining editability capabilities for further customizing processing.
    \item Extensive experiments demonstrate that PosterVerse can generate visually appealing posters with aesthetic designs, precise text, and well-crafted layouts, meeting the standards of commercial-grade posters.
\end{itemize}

\section{Related Work}
\subsubsection{Visual Text Image Synthesis}
In recent years, text-to-image generation has proliferated due to its unprecedented controllability and high fidelity \cite{ganfirstt2i2016icml,t2isurvey2024}. However, visual text synthesis remains challenging, requiring models to accurately render font structures while maintaining visual aesthetics. Early approaches improve text encoders by scaling them up \cite{balaji2022ediff,deepfloydif2023} or re-aligning them with visual features \cite{udifftext2024eccv}. Subsequently, for enhanced controllability and accuracy, researchers focus on conditioning diffusion models \cite{ddpm2020ho,ldm2022cvpr} with various prior information, which can be categorized into three types. The first type employs glyph images rendered on white backgrounds as conditions \cite{glyphdraw2023ma,glyphcontrol2023yang}. The second type combines a position mask and a rendered glyph (sometimes not) as input, using binary masks to refine text positions \cite{textdiffuser2023,anytext2,artist2025,brush2024,textflux2025xie}. In contrast to glyph or position masks, the third type extracts layout information for multiple text instances. TextDiffuser-2 \cite{textdiffuser22024} uses one LLM to generate language-like text layout and another to encode it as diffusion inputs. Lakhanpal et al. \cite{layout2025wacv} propose a training-free framework, using a frozen layout generator for iterative refinement.

\subsubsection{Layout Planning}
Layout planning is a crucial aspect to maintain natural text-background integration and visual aesthetics in text image synthesis. Preliminary studies focus on conditional layout generation using Transformer \cite{layouttf2021} and sequential Diffusion models \cite{ldgm2023}, producing bounding-box layouts without visual content. With the rise of LLM, their semantic and logical reasoning abilities are explored for layout planning \cite{layoutprompter2023,tang2024layoutnuwa,lggpt2025zhang}. Researchers then shift to content-aware layout generation, feeding background images and text prompts to models that place text appropriately without obscuring main content \cite{retrieval2024, Posterllama,postero2025hsu}. Recently, leveraging MLLMs' understanding capabilities, some works input background images and user requirements to generate typography JSON files that plan layouts with content-awareness and render textual content in tandem for better text arrangement and visual coherence \cite{posterllava,cole}. 

\subsubsection{Poster Generation}
Building upon advances in visual text synthesis and layout planning, automatic poster generation emerges as a specialized task that creates infographics with rich text and high-quality artistic presentation, primarily for advertising and marketing campaigns. Current poster generation methods can be grouped into two categories. The first category is semi-automatic, relying on pre-given conditions like positional masks \cite{anytext2}, user-specified subjects \cite{postermaker}, and text bounding boxes \cite{bizgen}. This heavy reliance on conditions greatly hampers their flexibility and user-friendliness. Conversely, the second type is fully-automatic and condition-free, leveraging only textual prompts to automatically plan layouts and generate visual content \cite{prompt2poster,postercraft,designdiffusion2025cvpr,posta}. This approach offers enhanced accessibility, enabling users to create commercial-grade posters through natural language descriptions. Despite the enhanced automation and visual quality, existing methods typically fall short in generating large-amount, high-density, and small-scale text, outputting illegible or misplaced characters. Also, they typically generate static posters, prohibiting post-editing. This motivates us to develop PosterVerse, a prompt-driven poster generation framework that novelly generates editable HTML-based typography file for poster design, enabling scalable text rendering and flexible post-generation customizability.

\section{Method}
\subsubsection{Overall Architecture}
The framework of PosterVerse is demonstrated in Fig.~\ref{fig: posterverse_pipeline}. It takes only the user requirement as input, and then designs a complete poster following three stages: blueprint creation, graphical background generation, and unified layout-text rendering. This architecture mirrors the workflow of professional designers while maintaining complete automation.

\subsubsection{Blueprint Creation}
User requirements for poster design are typically expressed through natural language, which tends to be ambiguous and lacks specificity. Designers should interpret these vague descriptions to understand the user's intentions, regardless of whether they are detailed or brief. Inspired by this, we design a Detail-Insensitive Requirement Parsing (DIPR) mechanism for the first stage of blueprint creation. We established three different user requirement levels (basic, medium, detailed) and trained the model to transform them into consistently comprehensive generation blueprints. The output blueprint is formatted as JSON and includes three parts: \textit{textual content} (e.g., title, subtitle, main content, and contact information), \textit{background graphical attributes} (e.g., style and image captions), and \textit{key extracted parameters} (e.g., resolution, theme, elements, color, and purpose). During DIPR training, we fine-tuned Qwen2.5-14B~\cite{qwen2.5} using randomly selected detail levels as input. The model is supervised by the same ground-truths of the blueprint information, thus developing insensitiveness to the richness of user input. The generated blueprints are used for training in subsequent stages.

\subsubsection{Graphical Background Generation}
The second stage of PosterVerse generates a graphical background for the poster. Background generation plays a crucial role in defining the overall aesthetic and tone of the poster. Motivated by professional designers who tailor their artistic styles to match project requirements, we classify poster backgrounds into four styles: \textit{Illustrative}, \textit{Design-Oriented}, \textit{Minimalistic}, and \textit{Photorealistic}. We fine-tuned Flux.1-dev~\cite{flux} using LoRA~\cite{lora} to obtain a specialized T2I model for each background type, respectively.

To further improve the quality of the output images, we integrated two core techniques into the training process. First, we implemented a \textit{resolution-based data bucketing strategy}, grouping training images by resolution and aspect ratio. This ensured that each batch contained visually consistent samples, preserving artistic composition and avoiding instability caused by mixing images with varying resolutions. 
Second, we introduced a \textit{dynamic prompt sampling mechanism}. Instead of using single prompts, we set up three hierarchical prompt levels, where basic prompts describe core visual elements and themes; medium prompts add artistic style; and detailed prompts precisely specify color, composition, and underlying meaning. Note that these three levels differ from those in the blueprint creation stage (the first stage) and are designed exclusively for the training of this stage. This hierarchical approach enables the generation model to adapt to diverse textual descriptions, significantly improving the diversity and accuracy of generated outputs. To train the four T2I models, we construct a tailored dataset that pairs hierarchical prompts and background images. For inference, the background graphical attributes generated from the first stage are fed into the model for background generation.

\subsubsection{Unified Layout-Text Rendering}
In the third stage, PosterVerse consolidates the foundational outputs from previous stages for complete poster generation, as depicted in the bottom left panel of Fig.~\ref{fig: posterverse_pipeline}. Unlike previous models that output static posters~\cite{posterllava, postermaker}, we innovatively choose HTML as the output format due to the following merits. (1) HTML's inherent text scalability perfectly addresses small-scale, high-density text synthesis that previous models struggle with. (2) HTML enables flexible post-editing of fonts, text, and layouts for frequently changing requirements; (3) HTML simultaneously covers layout and text rendering, avoiding the complexity of separate planning and generation. Specifically, we fine-tune Qwen2.5-VL-7B~\cite{qwen2.5-vl} to generate HTML files using textual content and parameters from the first stage plus background images from the second stage. The model is tasked with layout planning, graphical rendering, and text rendering as per the given inputs, producing a cohesive and aesthetically pleasing design. The dataset used for training is described in the next section. Finally, the generated HTML file can be rendered in a web browser, ensuring 100\% text fidelity and high-quality visual presentation. PosterVerse provides users with both editable HTML files and final image assets suitable for various distribution purposes.

\begin{figure}[t]
  \includegraphics[width=0.98\columnwidth]{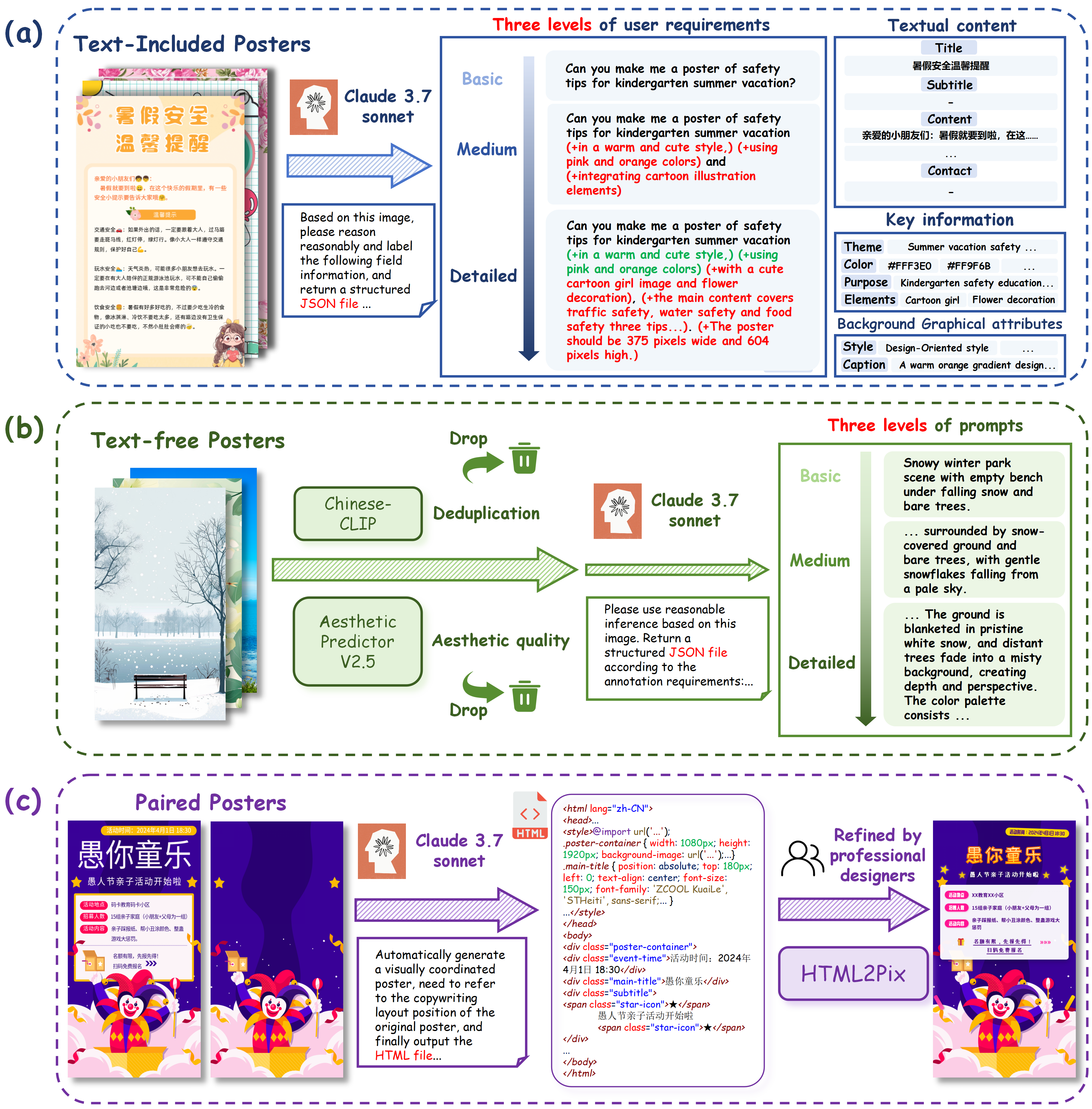}
  \centering
  \caption{Overview of the three core components in the PosterDNA dataset generation pipeline.}
  \label{fig: data_pipe}
\end{figure}

\section{PosterDNA Dataset}
\label{sec:dataset}

\begin{table*}[t]
\centering
\caption{Comparison of PosterVerse with existing methods. `Ave.', `PA.', `TA.', `IQ.', and `LC.' indicate Average, Prompt Adherence, Text Accuracy, Image Quality, and Layout \& Composition, respectively. {\textit{{\small{Open}}}} and {\textit{{\small{Close}}}} denote open-source and closed-source. The inputs for all methods are aligned with user requirements at the detailed level on the testing set.}
\begin{tabular}{lrrrrrcccccccc}
\hline
\multirow{2}{*}{Method}    & \multirow{2}{*}[0ex]{CR ↑}  & \multirow{2}{*}[0ex]{F1 ↑} & \multirow{2}{*}[0ex]{FID ↓} & \multirow{2}{*}[0ex]{Ove. ↓}  & \multicolumn{2}{c}{User Study ↑} & \multicolumn{1}{@{}c@{}}{} & \multicolumn{5}{c}{GPT-4o Evaluations ↑}     \\ \cline{6-7}  \cline{9-13} 
  & \multicolumn{1}{c}{}& \multicolumn{1}{c}{}  & \multicolumn{1}{c}{}     & \multicolumn{1}{c}{}   & \multicolumn{1}{c}{Vote}  & \multicolumn{1}{c}{Ave.} & \multicolumn{1}{c}{} & \multicolumn{1}{c}{Ave.} & \multicolumn{1}{c}{PA.} & \multicolumn{1}{c}{TA.} & \multicolumn{1}{c}{IQ.} & \multicolumn{1}{c}{LC.} \\ \hline
 \multicolumn{13}{c}{\textit{\textbf{Text-to-Image (T2I) Models}}}      \\ \hline
Kolors~{\textit{{\small{(Open)}}}}  & 4.59\% & 2.30\% & 123.41 & 0.0138 & 0\% & 2.26 & & 5.59 & 5.64 & 3.84 & 6.92 & 5.95   \\
Cogview4~{\textit{{\small{(Open)}}}} & 31.13\% & 21.01\% & 78.20 & 0.0140 & 0\% & 4.67 & & 6.03 & 6.42 & 5.56 & 5.81 & 6.33   \\
Ideogramv3~{\textit{{\small{(Close)}}}} & 30.57\% & 19.71\% & 97.40 & 0.0167 & 1\% & 4.81 & & 6.53 & 6.59 & 6.10 & 6.87 & 6.57   \\
Klingv2~{\textit{{\small{(Close)}}}}  & 35.27\% & 26.81\% & 71.64 & 0.0118 & 1\% & 5.13 & & 6.21 & 6.33 & 5.49 & 6.72 & 6.30    \\
Jimeng2.1~{\textit{{\small{(Close)}}}}  & 33.25\% & 23.01\% & \underline{68.70} & 0.0112 & 0\% & 5.53 & & 6.25 & 6.29 & 5.64 & 6.77 & 6.31   \\ \hline
 \multicolumn{13}{c}{\textit{\textbf{Unified Generative Models}}}  \\ \hline
Seedream3.0~{\textit{{\small{(Close)}}}}  & \underline{49.91\%} & 39.66\% & 83.20 & 0.0103 & 2\% & 6.12 & & 7.82 & \underline{7.99} & 7.55 & \textbf{8.03} & 7.71 \\
Gemini2.0~{\textit{{\small{(Close)}}}} & 38.22\% & 28.46\% & 74.39 & \underline{0.0086} & 1\% & 4.53 & & 6.22 & 6.49 & 5.69 & 6.53 & 6.18 \\
GPT-4o~{\textit{{\small{(Close)}}}} & 49.73\% & \underline{48.49\%} & 89.39 & 0.0106 & \underline{24\%} & \underline{6.30} & & \underline{7.87} & 7.93 & \underline{7.92} & \underline{7.89} & \textbf{7.73}    \\ \hline
\multicolumn{13}{c}{\textit{\textbf{Specialized Poster Generation Models}}}     \\ \hline
Anytext2~{\textit{{\small{(Open)}}}}   & 32.57\% & 26.46\% & 87.68 & 0.0105 & 0\% & 2.51 & & 3.78 & 3.78 & 3.30 & 4.27 & 3.77     \\
PosterMaker~{\textit{{\small{(Open)}}}} & 27.25\% & 25.09\% & 78.01 & 0.0098 & 0\%  & 3.08 & & 4.74 & 4.82 & 3.95 & 5.44 & 4.75   \\
Bizgen~{\textit{{\small{(Open)}}}}    & 14.67\% & 13.32\% & 101.86 & 0.0094 & 0\%  & 2.05 & & 3.46 & 3.06 & 3.19 & 4.25 & 3.34   \\ \hline
 PosterVerse (Ours)    & \textbf{92.33\%} & \textbf{78.58\%} & \textbf{62.54} & \textbf{0.0027} & \textbf{71\%} & \textbf{6.85} & & \textbf{8.02} & \textbf{8.19} & \textbf{8.51} & 7.66 & \underline{7.72}  \\ \hline
\end{tabular}
\label{tab:compare}
\end{table*}

Currently, while some layout generation datasets~\cite{CGL,posterlayout} have been published, they largely suffer from small scale and limited diversity. Also, none of them has explored a flexible, editable poster format that supports post-generation customization.
To fill this gap, we present PosterDNA, consisting of three specialized subsets: blueprint-creation (57,000 samples), graphic generation (100,000 samples), and unified text-layout creation (9,000 samples), responsible for PosterVerse's three-stage training and evaluation. An intuitive description of data construction is demonstrated in Fig.~\ref{fig: data_pipe}. We elaborate on the three training subsets, followed by the testing methodology.

\subsubsection{Blueprint Creation Subset}
The first stage of PosterVerse transforms user requirements into a consistently detailed blueprint regardless of the input's detail level. Hence, we curated 57,000 high-quality posters featuring rich, dense textual content to form the blueprint creation subset, as illustrated in the top panel of Fig.~\ref{fig: data_pipe}. To begin with, we employed Claude 3.7 Sonnet \cite{claude} to reverse-engineer three different detail levels (basic, medium, detailed) of user requirements used to generate these posters. Subsequently, we use Claude to extract refined requirement clues for each poster, including \textit{Key Information}, \textit{Background Graphical Attributes}, and \textit{Textual Content}. \textit{Key Information} includes poster's theme, dominant colors (hex codes), intended purpose (e.g., event promotion), and key visual elements (such as icons, objects, and decorations). \textit{Background Graphical Attributes} includes the poster's design style (e.g., Design-Oriented style) and graphical captions, in which the caption is the output portion that should be insensitive to the input requirements' detail levels. \textit{Textual Content} contains titles, subtitles, main body text, and contact information. During training, the three levels of user requirements are addressed to the Qwen2.5-14B for fine-tuning, with \textit{Textual Content}, \textit{Key Information}, and \textit{Background Graphical Attributes} as supervision labels.

\subsubsection{Graphic Generation Subset}
In the second stage, PosterVerse fine-tuned Flux.1-dev using LoRA to obtain four specialized instances for generating distinct background types. To support this training process, we constructed the graphic generation subset (middle of Fig.~\ref{fig: data_pipe}), constituting 100,000 text-free poster background graphics with diverse styles and designs. To retain only high-quality samples, we designed a multi-stage pipeline to verify image resolution and file format, as well as perform deduplication using a pretrained Chinese-CLIP \cite{Chinese-clip} model and aesthetic filtering with Aesthetic Predictor V2.5 \cite{aesthetic-predictor-v2-5}. Following quality filtering, we exploited Claude to generate prompts for each background, forming a prompt-image pair for model training. 
Corresponding to the dynamic caption sampling mechanism, we instruct Claude to generate three prompts with hierarchical details. Note that only one randomly selected prompt constitutes the input during training.

\subsubsection{Unified Layout-Text Rendering Subset}
The third stage of PosterVerse performs unified layout planning for visual elements and text while rendering typography. This stage requires the refined requirement specifications and a pre-given background image as input and delivers the HTML output file. Hence, we select 9,000 posters from the blueprint creation set to construct a unified text-layout rendering subset, as shown in the bottom panel of Fig.~\ref{fig: data_pipe}. We removed all text in the posters to obtain a text-free version, then paired the original text-included and text-free versions as input to Claude 3.7 Sonnet for generating corresponding HTML representations that capture the original layouts. Each HTML file underwent manual review by professional designers, including correcting positional errors and extracting text errors, adjusting element positions for better aesthetics, etc. These HTML files provide both structural graphical layouts and support diverse text rendering effects, enabling unified text-layout creation. For training, the text-free background along with Key Information and Textual Content extracted from the first stage are fed into the model, and the manually labeled HTML files serve as the supervision.

Additionally, we collected 1000 samples external to the training data to form the test set. To create the ground truth for each sample, they consistently went through the processing of the blueprint creation subset and unified layout-text rendering subsets. The ground-truths include ``basic-medium-detailed'' user requirements, key information, textual content, paired posters, and corresponding HTML files.

Constructing PosterDNA has consumed four months of manual effort, encompassing workflow design, data curation, and meticulous correction. Notably, the manual correction phase was the most intensive, accounting for approximately 80\% of the total labor. PosterDNA is the first Chinese poster generation dataset that pioneers in equipping HTML typography files for scalable text rendering, thus fundamentally addressing the challenge of small, large-amount, and high-density text rendering. It not only paves the way for commercial-grade poster design in text-rich cases but also contributes to the development of more dedicated methods.

\begin{figure*}[t]
  \includegraphics[width=1\linewidth]{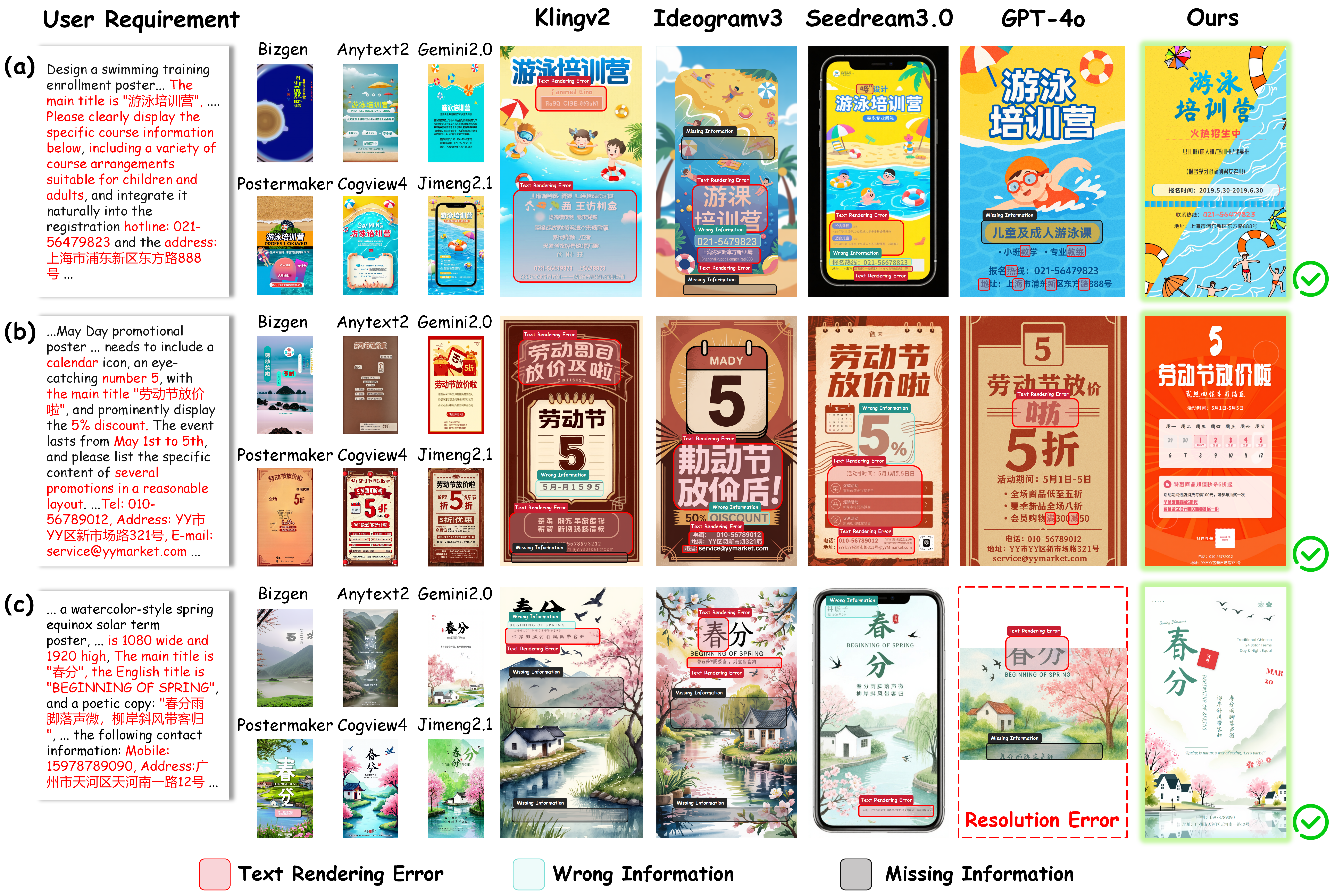}
  \centering
  \caption{Visual comparison of PosterVerse with state-of-the-art models. The inputs for all methods are aligned with user requirements at the detailed level on the testing set.}
  \label{fig: show_big}
\end{figure*}

\section{Experiments}
\subsection{Implementation Details}

\subsubsection{Model Implementation} 
For blueprint creation, we fine-tuned the Qwen-2.5-14B model using full-parameter SFT for 15 epochs at a 1e-5 learning rate on 8 H800 GPUs, completing in 30 hours. For graphical background generation, we fine-tuned Flux.1 dev with LoRA (rank 64) at a 5e-4 learning rate for 50 epochs. For unified layout-text rendering, we fine-tuned the Qwen2.5-VL-7B model at a 1e-5 learning rate for 50 epochs in 30 hours on 8 H800 GPUs. More details are included in the supplementary materials.

\subsubsection{Evaluation}
We assess our framework using objective quantitative metrics and subjective evaluation from both AI and human users. As for objective analysis, we measure text generation accuracy using Correct Rate (CR) and F1 scores (via PPOCRv5~\cite{ppocr}) following~\cite{hiercode}. We also assess layout fidelity with overlap metrics~\cite{posterlayout} and quantify perceptual similarity using FID~\cite{fid}. For subjective evaluation, we utilize GPT-4o to provide a detailed, four-dimensional rating (1-10) of prompt adherence, text accuracy, image quality, and layout \& composition. Furthermore, to evaluate real-world user experience, we conduct a human study where participants use the same four-dimensional rubric and also vote for their preferred model outputs. This dual approach to qualitative feedback ensures our evaluation is both comprehensive and truly reflective of the user experience.

\subsection{Comparison with Existing Methods}
We compare our method with 11 representative methods spanning three paradigms, including Text-to-Image models (Kolors~\cite{kolors}, Cogview4~\cite{cogview}, Ideogramv3~\cite{ideogram}, Klingv2~\cite{kling}, and Jimengv2,1~\cite{dreamposter}), unified generative models (Seedream3.0~\cite{seedream}, Gemini2.0-Flash-Gen~\cite{gemini}, and GPT-4o~\cite{gpt4ogen2025}), and specialized poster generation models (Anytext2~\cite{anytext2}, PosterMaker~\cite{postermaker}, and Bizgen~\cite{bizgen}). 

\subsubsection{Quantitative Comparison}
Quantitative comparison results between PosterVerse and baseline methods are presented in Tab.~\ref{tab:compare}, where consistent user requirements are used as inputs for fairness. On multiple objective metrics, PosterVerse achieves the best performance with a CR score of 92.33\% and an F1 score of 78.58\%, surpassing existing models by at least 42.42\% in CR and 30.09\% in F1 score. This not only demonstrates its effectiveness in producing accurate and readable text content but also highlights its ability to align closely with user requirements for copywriting. Furthermore, PosterVerse achieves an FID score of 62.54, highlighting the ability of PosterVerse to generate images with high perceptual similarity to poster visuals. In contrast, other models often exhibit text rendering errors and subpar background quality, resulting in lower FID scores. In addition, PosterVerse achieves the best Overlap score of 0.0027, reflecting its superior layout quality. 

\begin{figure}[t]
  \includegraphics[width=0.9\linewidth]{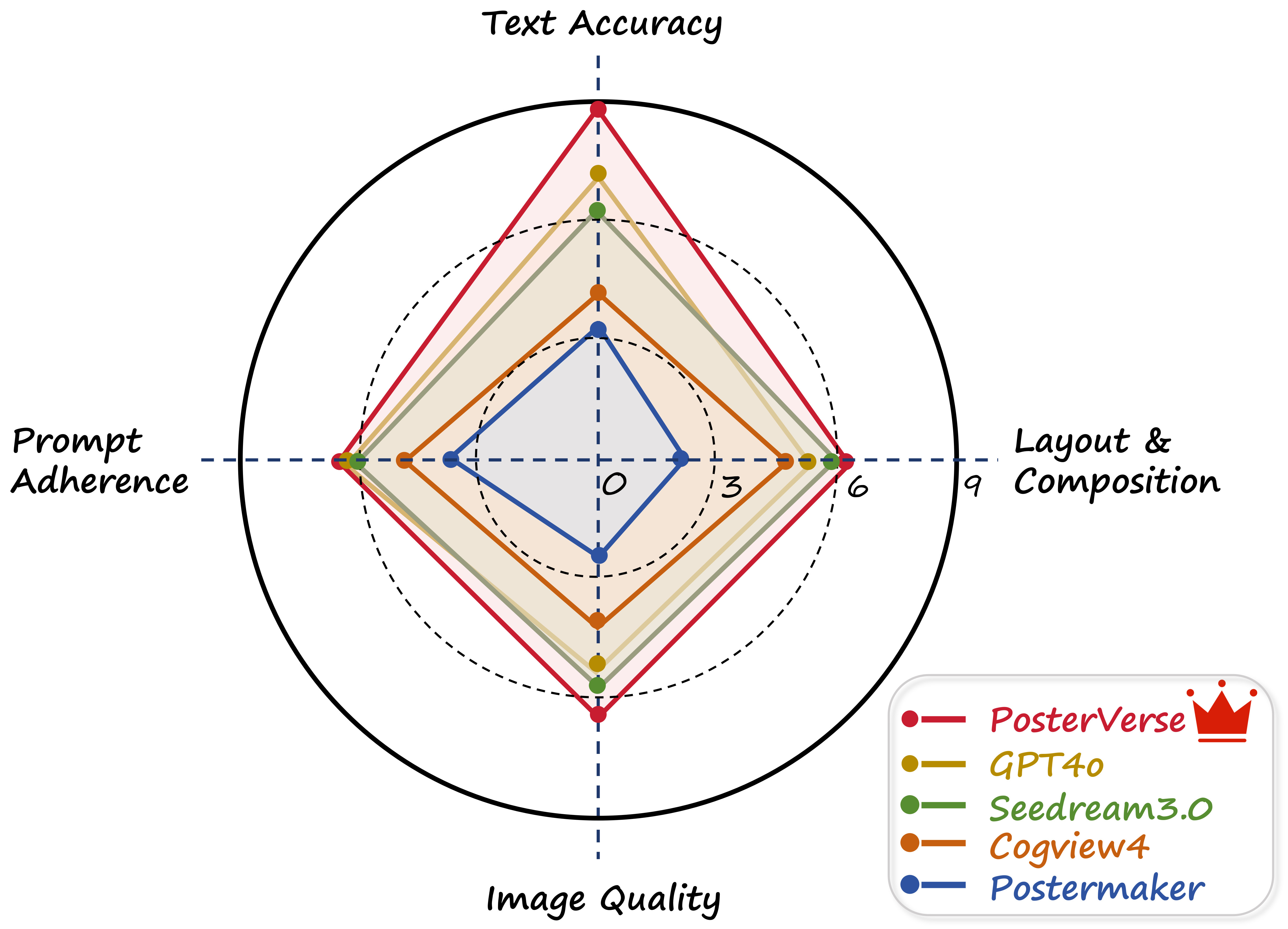}
  \centering
  \caption{Radar chart of the user study in four dimensions.}
  \label{fig: userstudy}
\end{figure}

Regarding subjective assessment, PosterVerse demonstrates exceptional performance across GPT-4o's four evaluation dimensions, achieving the highest overall average score, particularly excelling in Prompt Adherence and Text Accuracy. Furthermore, we invited 30 poster design users to conduct a user study comparing the poster generation performance of different models. As shown in Fig.~\ref{fig: userstudy}, partial results of the user study across the four key dimensions (aligned with the GPT-4o evaluation) are presented, with the corresponding average scores listed in the 7th column of Tab.~\ref{tab:compare}. PosterVerse demonstrated comparable performance to GPT-4o and Seedream3.0 across most dimensions, showing a significant advantage in textual accuracy. For the voting user study, users were asked to compare the posters generated by the models under the same input requirements and select the one with the best overall impression and practicality. As shown in the 6th column of Tab.~\ref{tab:compare}, PosterVerse received 71\% of the votes, significantly outperforming the other methods.

\subsubsection{Qualitative Comparison}
We present the visualizations of PosterVerse and the existing methods on the testing set. As illustrated in Fig.~\ref{fig: show_big}, we identify three main types of errors in existing methods. (1) The areas marked with red boxes highlight \textit{text rendering errors}. Models like Anytext2 and Cogview4 struggle to generate completely accurate text regardless of font size, while Seedream3.0 and GPT-4o perform better but still face significant challenges when handling dense text and small fonts. In contrast, PosterVerse is capable of accurate text rendering. (2) The areas marked with green boxes indicate instances where the rendered text contains \textit{wrong information}. For example, as shown in Fig.~\ref{fig: show_big} (a), while the user specified the phone number on the poster as ``021-56479823, Ideogramv3 incorrectly rendered it as ``021-5479823''. (3) The areas marked with gray boxes indicate cases where \textit{missing information} occurs. For instance, while the user requested the poster to include a phone number and physical address, Klingv2 missed rendering the required information. Moreover, GPT-4o often misunderstands user requirements regarding resolution, such as generating a landscape poster when a portrait poster was requested. In contrast, PosterVerse not only effectively aligns with user requirements but also supplements unclear user inputs, providing a more comprehensive and accurate output. 

Extended results under the English scenarios are included in the supplementary materials.

\begin{table}[]
\centering
\caption{Results of the user study demonstrating the effectiveness of the DIPR mechanism.}
\begin{tabular}{lcccc}
\hline
Method                             & Basic                & Medium               & Detailed             & \multicolumn{1}{c}{Ave. ↑} \\ \hline
\multirow{3}{*}{\thead{Seedream\\3.0}}       & \ding{51} &  &   & 4.92                         \\
                                   &  & \ding{51} &   & 5.33                 \\
                                   &  &   & \ding{51} & 6.12 \\ \hline
\multirow{3}{*}{GPT-4o}            &  \ding{51} &  &   & 4.52                         \\
                                   &    &  \ding{51}  &     & 6.45              \\
                                   & &  & \ding{51} & 6.30                         \\ \hline
\multirow{3}{*}{\thead{PosterVerse\\(Ours)}} & \ding{51} &  &  & \textbf{6.76}                         \\
                                   &  & \ding{51} &  & \textbf{6.53}                         \\
                                   &  &  & \ding{51} & \textbf{6.85}                         \\ \hline
\end{tabular}
\label{tab:requirement_ablation}
\end{table}

\begin{table}[]
\centering
\caption{Ablation study on the effectiveness of the dynamic prompt sampling mechanism.}
\begin{tabular}{ccccrr}
\hline
\#Line & Basic & Medium & Detailed & \multicolumn{1}{c}{FID ↓} & \multicolumn{1}{c}{IS ↑} \\ \hline
1     & \ding{55} &  \ding{55}  &  \ding{51}   & 136.72  &  62.39 \\
2     &  \ding{51} &  \ding{51}  &  \ding{51} & \textbf{62.54} & \textbf{77.85} \\ \hline
\end{tabular}
\label{tab:caption_ablation}
\end{table}

\subsection{Ablation Study}
We conducted an ablation study through human evaluation (with the same setting as before) to investigate DIPR mechanism's effectiveness. As shown in Tab.~\ref{tab:requirement_ablation}, we observe that the two representative baseline models are highly sensitive to input detail levels. When given basic requirements, their average ratings are low, indicating their deficiency in generating high-quality posters with brief requirements. In contrast, PosterVerse maintains consistently high performance across all input detail levels, validating DIPR's effectiveness.

Additionally, to further validate that the dynamic prompt sampling mechanism during the second stage of PosterVerse training can enhance the effectiveness of graphical background generation, we conducted a comparison with models trained using only the detailed-level prompt. As shown in Tab.~\ref{tab:caption_ablation}, both the FID and CLIP-IS metrics are significantly better when using hierarchical prompts compared to using only the detailed-level prompt.

\section{Conclusion}
In this paper, we present PosterVerse, a full-workflow method that seamlessly combines blueprint creation, graphical background generation, and unified layout-text rendering, enabling commercial-grade posters with sophisticated layouts and text-dense designs. Additionally, we introduce PosterDNA, the first high-quality, text-dense poster generation dataset with fine-grained HTML-based specifications, tailored for modular training and validation. Extensive experiments demonstrate PosterVerse's superior performance, significantly outperforming existing methods. The PosterVerse's ability to generate commercial-grade posters directly from natural language prompts, combined with its scalable and editable output format, establishes a new paradigm for automated commercial design and provides a promising solution for marketing and creative industries.

\section{Limitation}
Despite its strengths, PosterVerse has certain limitations. Due to its workflow-based approach, generating a single poster takes 2-3 minutes, which can be relatively time-consuming. Additionally, the multi-stage process introduces potential challenges, such as occasional misalignment between the graphical background and text layout, particularly in more complex or creative design scenarios.

\section{Acknowledgements}
This research is supported in part by the National Natural Science Foundation of China (Grant No.:62476093).

\bibliography{aaai2026}

@misc{posterllava,
      title={{PosterLLaVa: Constructing a Unified Multi-modal Layout Generator with LLM}}, 
      author={Tao Yang and Yingmin Luo and Zhongang Qi and Yang Wu and Ying Shan and Chang Wen Chen},
      year={2024},
      eprint={2406.02884},
      archivePrefix={arXiv},
      primaryClass={cs.CV},
}

@article{cole,
  title={{Cole: A Hierarchical Generation Framework for Multi-Layered and Editable Graphic Design}},
  author={Jia, Peidong and Li, Chenxuan and Yuan, Yuhui and Liu, Zeyu and Shen, Yichao and Chen, Bohan and Chen, Xingru and Zheng, Yinglin and Chen, Dong and Li, Ji and others},
  journal={arXiv preprint arXiv:2311.16974},
  year={2023}
}

@misc{flux,
  author = {{Black Forest Labs}},
  title = {Flux},
  year = {2024},
  url = {https://github.com/black-forest-labs/flux},
  note = {https://github.com/black-forest-labs/flux}
}

@InProceedings{Posterllama,
author="Seol, Jaejung
and Kim, Seojun
and Yoo, Jaejun",
title="PosterLlama: Bridging Design Ability of Language Model to Content-Aware Layout Generation",
booktitle="European Conference on Computer Vision (ECCV)",
year="2024",
pages="451--468",
}

@misc{aesthetic-predictor-v2-5,
  author = {{discus0434}},
  title = {{Aesthetic-Predictor-V2-5}},
  year = {2024},
  url = {https://github.com/discus0434/aesthetic-predictor-v2-5},
  note = {https://github.com/discus0434/aesthetic-predictor-v2-5}
}

@misc{sd3.5,
  author = {{Stability AI}},
  title = {{Stable Diffusion 3.5}},
  year = {2024},
  url = {https://github.com/Stability-AI/sd3.5},
  note = {https://github.com/Stability-AI/sd3.5}
}

@inproceedings{postermaker,
  title={{Postermaker: Towards high-quality product poster generation with accurate text rendering}},
  author={Gao, Yifan and Lin, Zihang and Liu, Chuanbin and Zhou, Min and Ge, Tiezheng and Zheng, Bo and Xie, Hongtao},
  booktitle={Proceedings of the Computer Vision and Pattern Recognition Conference (CVPR)},
  pages={8083--8093},
  year={2025}
}

@article{kolors,
  title={{Kolors: Effective Training of Diffusion Model for Photorealistic Text-to-Image Synthesis}},
  author={{Team, Kolors}},
  journal={arXiv preprint},
  year={2024}
}

@inproceedings{cogview,
  title={{Cogview3: Finer and Faster Text-to-Image Generation via Relay Diffusion}},
  author={Zheng, Wendi and Teng, Jiayan and Yang, Zhuoyi and Wang, Weihan and Chen, Jidong and Gu, Xiaotao and Dong, Yuxiao and Ding, Ming and Tang, Jie},
  booktitle={European Conference on Computer Vision (ECCV)},
  pages={1--22},
  year={2024},
  organization={Springer}
}

@inproceedings{bizgen,
  title={{Bizgen: Advancing Article-Level Visual Text Rendering for Infographics Generation}},
  author={Peng, Yuyang and Xiao, Shishi and Wu, Keming and Liao, Qisheng and Chen, Bohan and Lin, Kevin and Huang, Danqing and Li, Ji and Yuan, Yuhui},
  booktitle={Proceedings of the Computer Vision and Pattern Recognition Conference (CVPR)},
  pages={23615--23624},
  year={2025}
}

@misc{ideogram,
  author = {{Ideogram AI}},
  title = {{Ideogram v3}},
  year = {2025},
  url = {https://ideogram.ai/launch},
  note = {https://ideogram.ai/launch}
}

@article{gemini,
  title={{Gemini: A Family of Highly Capable Multimodal Models}},
  author={Gemini, Team and Anil, Rohan and Borgeaud, Sebastian and Alayrac, Jean-Baptiste and Yu, Jiahui and Soricut, Radu and Schalkwyk, Johan and Dai, Andrew M and Hauth, Anja and Millican, Katie and others},
  journal={arXiv preprint arXiv:2312.11805},
  year={2023}
}

@misc{gpt4ogen2025,
	author       = {OpenAI},
	title        = {Introducing GPT-4o Image Generation},
	howpublished = {\url{https://openai.com/index/introducing-4o-image-generation/}},
	year         = {2025},
}

@article{seedream,
  title={{Seedream 3.0 Technical Report}},
  author={Gao, Yu and Gong, Lixue and Guo, Qiushan and Hou, Xiaoxia and Lai, Zhichao and Li, Fanshi and Li, Liang and Lian, Xiaochen and Liao, Chao and Liu, Liyang and others},
  journal={arXiv preprint arXiv:2504.11346},
  year={2025}
}

@article{postercraft,
  title={{PosterCraft: Rethinking High-Quality Aesthetic Poster Generation in a Unified Framework}},
  author={Chen, SiXiang and Lai, Jianyu and Gao, Jialin and Ye, Tian and Chen, Haoyu and Shi, Hengyu and Shao, Shitong and Lin, Yunlong and Fei, Song and Xing, Zhaohu and others},
  journal={arXiv preprint arXiv:2506.10741},
  year={2025}
}

@article{dreamposter,
  title={{DreamPoster: A Unified Framework for Image-Conditioned Generative Poster Design}},
  author={Hu, Xiwei and Chen, Haokun and Qi, Zhongqi and Zhang, Hui and Hong, Dexiang and Shao, Jie and Wu, Xinglong},
  journal={arXiv preprint arXiv:2507.04218},
  year={2025}
}

@inproceedings{posta,
  title={{POSTA: A Go-to Framework for Customized Artistic Poster Generation}},
  author={Chen, Haoyu and Xu, Xiaojie and Li, Wenbo and Ren, Jingjing and Ye, Tian and Liu, Songhua and Chen, Ying-Cong and Zhu, Lei and Wang, Xinchao},
  booktitle={Proceedings of the Computer Vision and Pattern Recognition Conference (CVPR)},
  pages={28694--28704},
  year={2025}
}

@inproceedings{Relation-Aware-Diffusion-Model,
  title={{Relation-aware diffusion model for controllable poster layout generation}},
  author={Li, Fengheng and Liu, An and Feng, Wei and Zhu, Honghe and Li, Yaoyu and Zhang, Zheng and Lv, Jingjing and Zhu, Xin and Shen, Junjie and Lin, Zhangang and others},
  booktitle={Proceedings of the 32nd ACM International Conference on Information and Knowledge Management (CIKM)},
  pages={1249--1258},
  year={2023}
}

@article{Planning-and-rendering,
  title={{Planning and Rendering: Towards Product Poster Generation with Diffusion Models}},
  author={Li, Zhaochen and Li, Fengheng and Feng, Wei and Zhu, Honghe and Li, Yaoyu and Zhang, Zheng and Lv, Jingjing and Shen, Junjie and Lin, Zhangang and Shao, Jingping and others},
  journal={arXiv preprint arXiv:2312.08822},
  year={2023}
}

@inproceedings{posterlayout,
  title={{Posterlayout: A New Benchmark and Approach for Content-Aware Visual-Textual Presentation Layout}},
  author={Hsu, Hsiao Yuan and He, Xiangteng and Peng, Yuxin and Kong, Hao and Zhang, Qing},
  booktitle={Proceedings of the IEEE/CVF Conference on Computer Vision and Pattern Recognition (CVPR)},
  pages={6018--6026},
  year={2023}
}

@inproceedings{prompt2poster,
  title={{Prompt2poster: Automatically Artistic Chinese Poster Creation from Prompt Only}},
  author={Wang, Shaodong and Ge, Yunyang and Chen, Liuhan and Zhou, Haiyang and Wang, Qian and Cheng, Xinhua and Yuan, Li},
  booktitle={Proceedings of the 32nd ACM International Conference on Multimedia (MM)},
  pages={10716--10724},
  year={2024}
}

@inproceedings{autoposter,
  title={{AutoPoster: A highly Automatic and Content-Aware Design System for Advertising Poster Generation}},
  author={Lin, Jinpeng and Zhou, Min and Ma, Ye and Gao, Yifan and Fei, Chenxi and Chen, Yangjian and Yu, Zhang and Ge, Tiezheng},
  booktitle={Proceedings of the 31st ACM International Conference on Multimedia (MM)},
  pages={1250--1260},
  year={2023}
}

@article{qwen2.5-vl,
  title={{Qwen2.5-VL Technical Report}},
  author={Bai, Shuai and Chen, Keqin and Liu, Xuejing and Wang, Jialin and Ge, Wenbin and Song, Sibo and Dang, Kai and Wang, Peng and Wang, Shijie and Tang, Jun and others},
  journal={arXiv preprint arXiv:2502.13923},
  year={2025}
}

@article{qwen2.5,
    title={{Qwen2.5 Technical Report}},
    author={Yang, An and Yang, Baosong and Zhang, Beichen and Hui, Binyuan and Zheng, Bo and Yu, Bowen and Li, Chengyuan and Liu, Dayiheng and Huang, Fei and Wei, Haoran and others},
    journal={arXiv preprint arXiv:2412.15115},
    year={2024}
}

@article{Chinese-clip,
  title={{Chinese CLIP: Contrastive Vision-Language Pretraining in Chinese}},
  author={Yang, An and Pan, Junshu and Lin, Junyang and Men, Rui and Zhang, Yichang and Zhou, Jingren and Zhou, Chang},
  journal={arXiv preprint arXiv:2211.01335},
  year={2022}
}

@misc{claude,
  author = {{Anthropic}},
  title = {{Claude Sonnet}},
  year = {2024},
  url = {https://www.anthropic.com/claude/sonnet},
  note = {https://www.anthropic.com/claude/sonnet}
}

@article{lora,
  title={{LoRA: Low-Rank Adaptation of Large Language Models}},
  author={Hu, Edward J and Shen, Yelong and Wallis, Phillip and Allen-Zhu, Zeyuan and Li, Yuanzhi and Wang, Shean and Wang, Lu and Chen, Weizhu and others},
  journal={International Conference on Learning Representations (ICLR)},
  volume={1},
  number={2},
  pages={3},
  year={2022}
}

@article{anytext2,
  title={{Anytext2: Visual Text Generation and Editing with Customizable Attributes}},
  author={Tuo, Yuxiang and Geng, Yifeng and Bo, Liefeng},
  journal={arXiv preprint arXiv:2411.15245},
  year={2024}
}

@article{lggpt2025zhang,
  title={{Smaller But Better: Unifying Layout Generation with Smaller Large Language Models}},
  author={Zhang, Peirong and Zhang, Jiaxin and Cao, Jiahuan and Li, Hongliang and Jin, Lianwen},
  journal={International Journal of Computer Vision (IJCV)},
  volume={133},
  pages={3891–3917},
  year={2025}
}

@article{glyphdraw2023ma,
  title={{Glyphdraw: Seamlessly Rendering Text with Intricate Spatial Structures in Text-to-Image Generation}},
  author={Ma, Jian and Zhao, Mingjun and Chen, Chen and Wang, Ruichen and Niu, Di and Lu, Haonan and Lin, Xiaodong},
  journal={arXiv preprint arXiv:2303.17870},
  year={2023}
}

@inproceedings{glyphcontrol2023yang,
 author = {Yang, Yukang and Gui, Dongnan and YUAN, YUHUI and Liang, Weicong and Ding, Haisong and Hu, Han and Chen, Kai},
 booktitle = {Advances in Neural Information Processing Systems (NeurIPS)},
 pages = {44050--44066},
 title = {GlyphControl: Glyph Conditional Control for Visual Text Generation},
 volume = {36},
 year = {2023}
}

@article{textflux2025xie,
  title={TextFlux: An OCR-Free DiT Model for High-Fidelity Multilingual Scene Text Synthesis},
  author={Xie, Yu and Zhang, Jielei and Chen, Pengyu and Wang, Ziyue and Wang, Weihang and Gao, Longwen and Li, Peiyi and Sun, Huyang and Zhang, Qiang and Qiao, Qian and others},
  journal={arXiv preprint arXiv:2505.17778},
  year={2025}
}

@misc{deepfloydif2023,
    title={{DeepFloyd-IF}},
    author={DeepFloyd Lab},
    url={https://github.com/deep-floyd/if},
    year={2023}
}

@article{balaji2022ediff,
  title={ediff-i: Text-to-Image Diffusion Models with an Ensemble of Expert Denoisers},
  author={Balaji, Yogesh and Nah, Seungjun and Huang, Xun and Vahdat, Arash and Song, Jiaming and Zhang, Qinsheng and Kreis, Karsten and Aittala, Miika and Aila, Timo and Laine, Samuli and others},
  journal={arXiv preprint arXiv:2211.01324},
  year={2022}
}

@article{t2isurvey2024,
	title={Text-to-Image Synthesis: A Decade Survey}, 
	author={Nonghai Zhang, Hao Tang},
	year={2024},
	journal={arXiv preprint arXiv:2411.16164},
}

@InProceedings{ganfirstt2i2016icml,
	title = 	 {Generative Adversarial Text to Image Synthesis},
	author = 	 {Reed, Scott and Akata, Zeynep and Yan, Xinchen and Logeswaran, Lajanugen and Schiele, Bernt and Lee, Honglak},
	booktitle = 	 {Proceedings of The 33rd International Conference on Machine Learning (ICML)},
	pages = 	 {1060--1069},
	year = 	 {2016},
	volume = 	 {48},
	month = 	 {20--22 Jun},
}

@InProceedings{ldm2022cvpr,
	author    = {Rombach, Robin and Blattmann, Andreas and Lorenz, Dominik and Esser, Patrick and Ommer, Bj\"orn},
	title     = {High-Resolution Image Synthesis With Latent Diffusion Models},
	booktitle = {Proceedings of the IEEE/CVF Conference on Computer Vision and Pattern Recognition (CVPR)},
	month     = {June},
	year      = {2022},
	pages     = {10684-10695}
}

@InProceedings{udifftext2024eccv,
	author="Zhao, Yiming
	and Lian, Zhouhui",
	title={{UDiffText: A Unified Framework for High-Quality Text Synthesis in Arbitrary Images via Character-Aware Diffusion Models}},
	booktitle={European Conference on Computer Vision (ECCV)},
	year="2024",
	pages="217--233",
}

@inproceedings{ddpm2020ho,
	author = {Ho, Jonathan and Jain, Ajay and Abbeel, Pieter},
	booktitle = {Advances in Neural Information Processing Systems (NeurIPS)},
	pages = {6840--6851},
	title = {{Denoising Diffusion Probabilistic Models}},
	volume = {33},
	year = {2020}
}

@INPROCEEDINGS{layout2025wacv,
  author={Lakhanpal, Sanyam and Chopra, Shivang and Jain, Vinija and Chadha, Aman and Luo, Man},
  booktitle={IEEE/CVF Winter Conference on Applications of Computer Vision (WACV)}, 
  title={Refining Text-to-Image Generation: Towards Accurate Training-Free Glyph-Enhanced Image Generation}, 
  year={2025},
  volume={},
  number={},
  pages={4372-4381},
}

@INPROCEEDINGS{artist2025,
  author={Zhang, Jianyi and Zhou, Yufan and Gu, Jiuxiang and Wigington, Curtis and Yu, Tong and Chen, Yiran and Sun, Tong and Zhang, Ruiyi},
  booktitle={IEEE/CVF Winter Conference on Applications of Computer Vision (WACV)}, 
  title={ARTIST: Improving the Generation of Text-Rich Images with Disentangled Diffusion Models and Large Language Models}, 
  year={2025},
  volume={},
  number={},
  pages={1268-1278},
}

@article{brush2024, 
    title={Brush Your Text: Synthesize Any Scene Text on Images via Diffusion Model}, 
    volume={38}, 
    number={7},
    journal={Proceedings of the AAAI Conference on Artificial Intelligence},
    author={Zhang, Lingjun and Chen, Xinyuan and Wang, Yaohui and Lu, Yue and Qiao, Yu}, 
    year={2024}, 
    month={Mar.}, 
    pages={7215-7223}
}

@inproceedings{textdiffuser2023,
    author = {Chen, Jingye and Huang, Yupan and Lv, Tengchao and Cui, Lei and Chen, Qifeng and Wei, Furu},
    booktitle = {Advances in Neural Information Processing Systems (NeurIPS)},
    pages = {9353--9387},
    title = {TextDiffuser: Diffusion Models as Text Painters},
    volume = {36},
    year = {2023}
}

@InProceedings{textdiffuser22024,
author="Chen, Jingye
and Huang, Yupan
and Lv, Tengchao
and Cui, Lei
and Chen, Qifeng
and Wei, Furu",
title="TextDiffuser-2: Unleashing the Power of Language Models for Text Rendering",
booktitle="European Conference on Computer Vision (ECCV)",
year="2024",
address="Cham",
pages="386--402",
}

@InProceedings{layouttf2021,
    author    = {Gupta, Kamal and Lazarow, Justin and Achille, Alessandro and Davis, Larry S. and Mahadevan, Vijay and Shrivastava, Abhinav},
    title     = {LayoutTransformer: Layout Generation and Completion With Self-Attention},
    booktitle = {Proceedings of the IEEE/CVF International Conference on Computer Vision (ICCV)},
    month     = {October},
    year      = {2021},
    pages     = {1004-1014}
}

@InProceedings{ldgm2023,
    author    = {Hui, Mude and Zhang, Zhizheng and Zhang, Xiaoyi and Xie, Wenxuan and Wang, Yuwang and Lu, Yan},
    title     = {Unifying Layout Generation With a Decoupled Diffusion Model},
    booktitle = {Proceedings of the IEEE/CVF Conference on Computer Vision and Pattern Recognition (CVPR)},
    month     = {June},
    year      = {2023},
    pages     = {1942-1951}
}

@inproceedings{layoutprompter2023,
    author = {Lin, Jiawei and Guo, Jiaqi and Sun, Shizhao and Yang, Zijiang and Lou, Jian-Guang and Zhang, Dongmei},
    booktitle = {Advances in Neural Information Processing Systems (NeurIPS)},
    pages = {43852--43879},
    title = {LayoutPrompter: Awaken the Design Ability of Large Language Models},
    volume = {36},
    year = {2023}
}

@inproceedings{tang2024layoutnuwa,
    title={Layout{NUWA}: Revealing the Hidden Layout Expertise of Large Language Models},
    author={Zecheng Tang and Chenfei Wu and Juntao Li and Nan Duan},
    booktitle={International Conference on Learning Representations (ICLR)},
    year={2024},
}

@InProceedings{postero2025hsu,
    author    = {Hsu, Hsiao Yuan and Peng, Yuxin},
    title     = {{PosterO: Structuring Layout Trees to Enable Language Models in Generalized Content-Aware Layout Generation}},
    booktitle = {Proceedings of the IEEE/CVF Conference on Computer Vision and Pattern Recognition (CVPR)},
    month     = {June},
    year      = {2025},
    pages     = {8117-8127}
}

@InProceedings{retrieval2024,
    author    = {Horita, Daichi and Inoue, Naoto and Kikuchi, Kotaro and Yamaguchi, Kota and Aizawa, Kiyoharu},
    title     = {{Retrieval-Augmented Layout Transformer for Content-Aware Layout Generation}},
    booktitle = {Proceedings of the IEEE/CVF Conference on Computer Vision and Pattern Recognition (CVPR)},
    month     = {June},
    year      = {2024},
    pages     = {67-76}
}

@misc{ppocr,
      title={{PaddleOCR 3.0 Technical Report}}, 
      author={Cheng Cui and Ting Sun and Manhui Lin and Tingquan Gao and Yubo Zhang and Jiaxuan Liu and Xueqing Wang and Zelun Zhang and Changda Zhou and Hongen Liu and Yue Zhang and Wenyu Lv and Kui Huang and Yichao Zhang and Jing Zhang and Jun Zhang and Yi Liu and Dianhai Yu and Yanjun Ma},
      year={2025},
      eprint={2507.05595},
      archivePrefix={arXiv},
      primaryClass={cs.CV},
}

@article{hiercode,
  title={{Hiercode: A lightweight hierarchical codebook for zero-shot Chinese text recognition}},
  author={Zhang, Yuyi and Zhu, Yuanzhi and Peng, Dezhi and Zhang, Peirong and Yang, Zhenhua and Yang, Zhibo and Yao, Cong and Jin, Lianwen},
  journal={Pattern Recognition},
  volume={158},
  pages={110963},
  year={2025},
  publisher={Elsevier}
}

@article{fid,
  title={{GANs Trained by a Two Time-Scale Update Rule Converge to a Local Nash Equilibrium}},
  author={Heusel, Martin and Ramsauer, Hubert and Unterthiner, Thomas and Nessler, Bernhard and Hochreiter, Sepp},
  journal={Advances in Neural Information Processing Systems (NeurIPS)},
  volume={30},
  year={2017}
}

@misc{kling,
    title={{Klingv2}},
    author = {{Kling AI}},
    url={https://www.aigc.cn/kling-ai},
    year={2025}
}

@InProceedings{designdiffusion2025cvpr,
    author    = {Wang, Zhendong and Bao, Jianmin and Gu, Shuyang and Chen, Dong and Zhou, Wengang and Li, Houqiang},
    title     = {{DesignDiffusion: High-Quality Text-to-Design Image Generation with Diffusion Models}},
    booktitle = {Proceedings of the IEEE/CVF Conference on Computer Vision and Pattern Recognition (CVPR)},
    month     = {June},
    year      = {2025},
    pages     = {20906-20915}
}

@article{CGL,
  title={{Composition-Aware Graphic Layout GAN for Visual-Textual Presentation Designs}},
  author={Zhou, Min and Xu, Chenchen and Ma, Ye and Ge, Tiezheng and Jiang, Yuning and Xu, Weiwei},
  journal={arXiv preprint arXiv:2205.00303},
  year={2022}
}

@article{t5,
  title={{Exploring the Limits of Transfer Learning with a Unified Text-to-Text Transformer}},
  author={Raffel, Colin and Shazeer, Noam and Roberts, Adam and Lee, Katherine and Narang, Sharan and Matena, Michael and Zhou, Yanqi and Li, Wei and Liu, Peter J},
  journal={Journal of machine learning research},
  volume={21},
  number={140},
  pages={1--67},
  year={2020}
}

@inproceedings{layoutdetr,
  title={{Layoutdetr: detection transformer is a good multimodal layout designer}},
  author={Yu, Ning and Chen, Chia-Chih and Chen, Zeyuan and Meng, Rui and Wu, Gang and Josel, Paul and Niebles, Juan Carlos and Xiong, Caiming and Xu, Ran},
  booktitle={European Conference on Computer Vision},
  pages={169--187},
  year={2024},
  organization={Springer}
}

@inproceedings{youtube,
  title={Toward human perception-centric video thumbnail generation},
  author={Yang, Tao and Wang, Fan and Lin, Junfan and Qi, Zhongang and Wu, Yang and Xu, Jing and Shan, Ying and Chen, Changwen},
  booktitle={Proceedings of the 31st ACM International Conference on Multimedia},
  pages={6653--6664},
  year={2023}
}

\clearpage
\twocolumn[
  \begin{@twocolumnfalse}
    \begin{center}
      {\LARGE\bfseries PosterVerse: A Full-Workflow Framework for Commercial-Grade Poster Generation with HTML-Based Scalable Typography\\Supplementary Material\par}
      \vspace{0.8em}
    \end{center}
  \end{@twocolumnfalse}
]

\appendix 

\begin{figure*}[t]
  \includegraphics[width=1\linewidth]{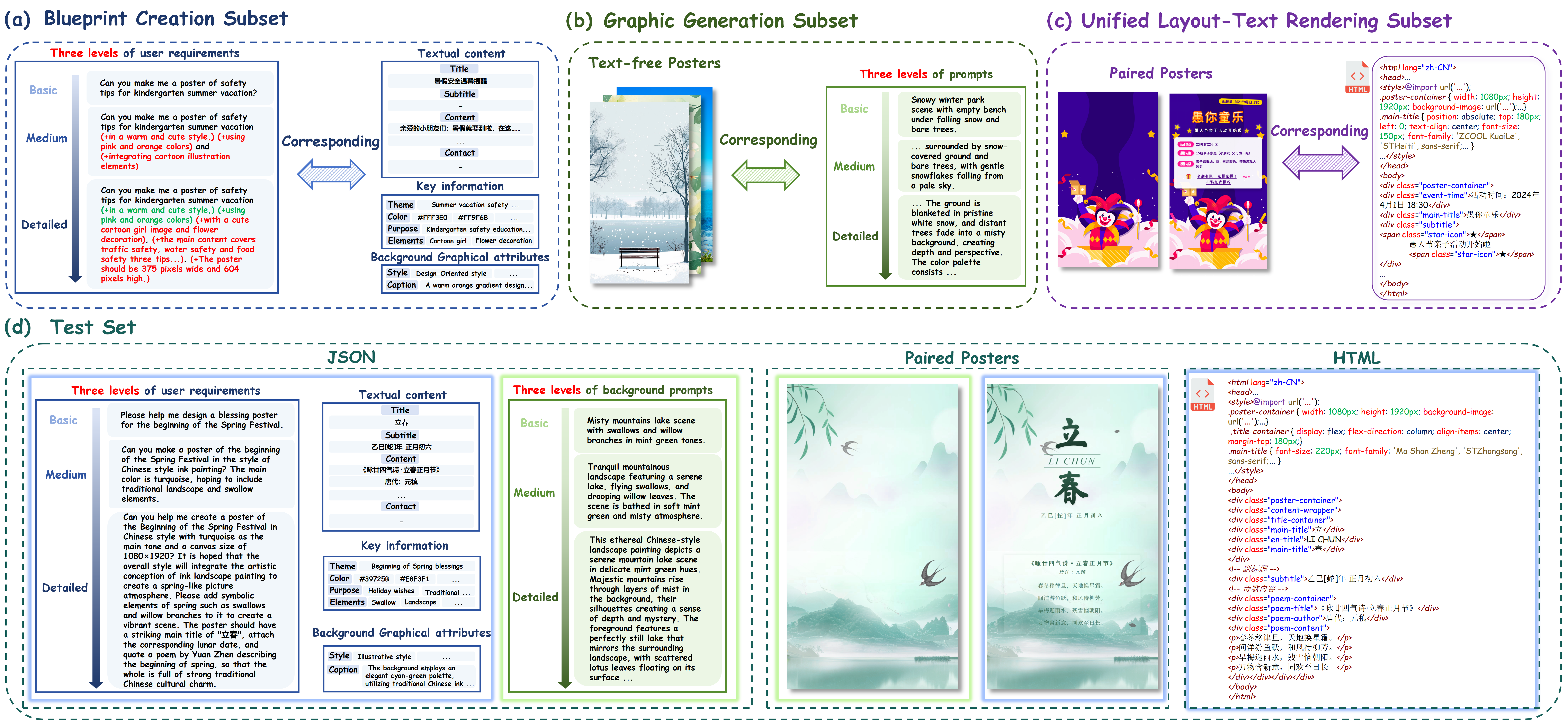}
  \centering
  \caption{The Visualization of the PosterDNA Dataset.}
  \label{fig: dataset}
\end{figure*}

\section{More Details of PosterDNA Dataset}
\subsection{Dataset Structure}
\begin{figure*}[t]
  \includegraphics[width=1\linewidth]{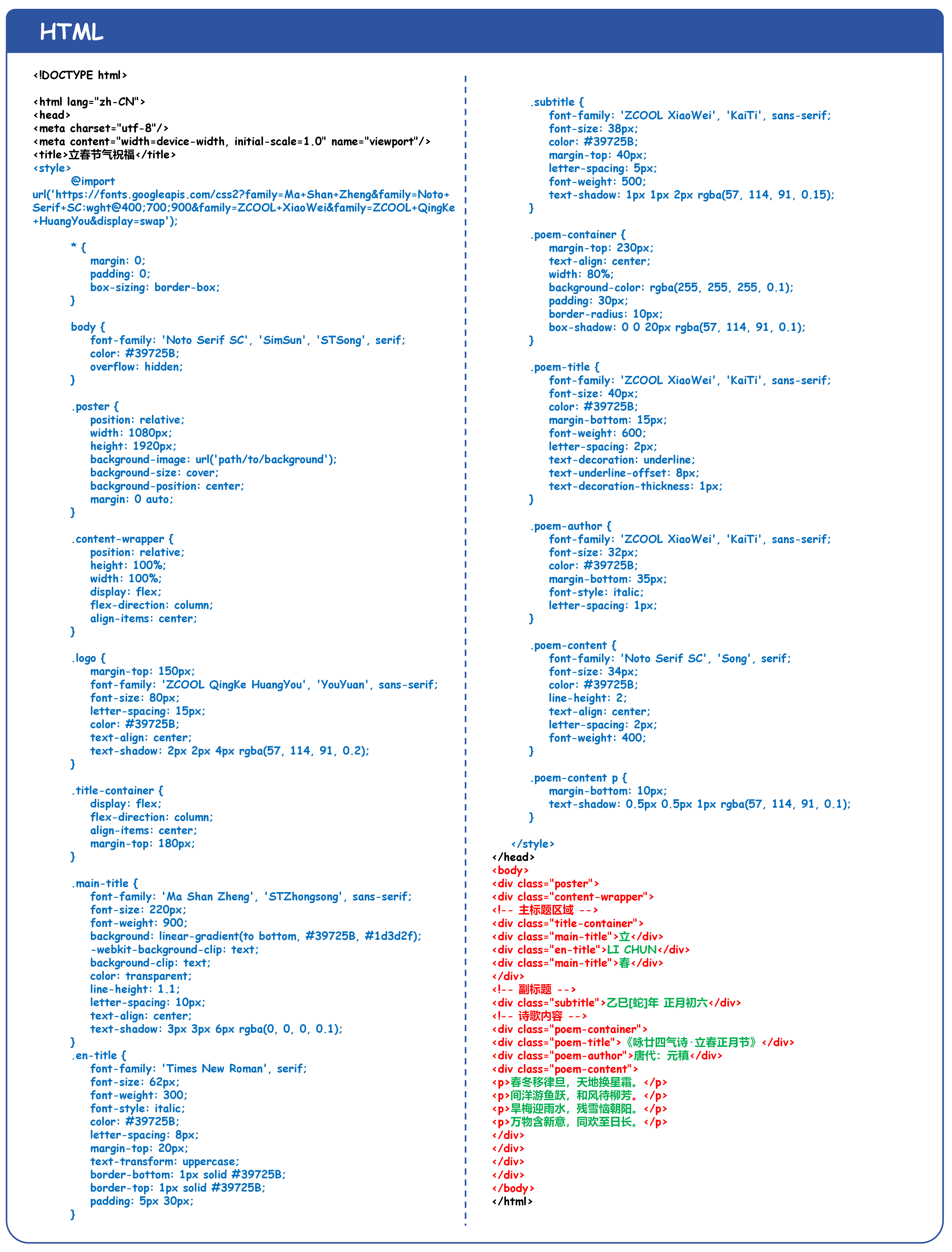}
  \centering
  \caption{An example of the unified layout-text annotation (HTML format).}
  \label{fig: html}
\end{figure*}

PosterDNA consists of three specialized subsets: blueprint creation (57,000 samples), graphic generation (100,000 samples), and unified text-layout creation (9,000 samples). Additionally, PosterDNA includes a test set of 1,000 samples. The test set contains "basic-medium-detailed" user requirements, key information, and textual content (in JSON format), paired posters, and corresponding unified layout-text annotations (in HTML format). A visualization of the dataset structure is shown in Fig.\ref{fig: dataset}, and an example of the unified layout-text annotation (HTML format) is presented in Fig.\ref{fig: html}.

\begin{figure*}[t]
  \includegraphics[width=0.75\linewidth]{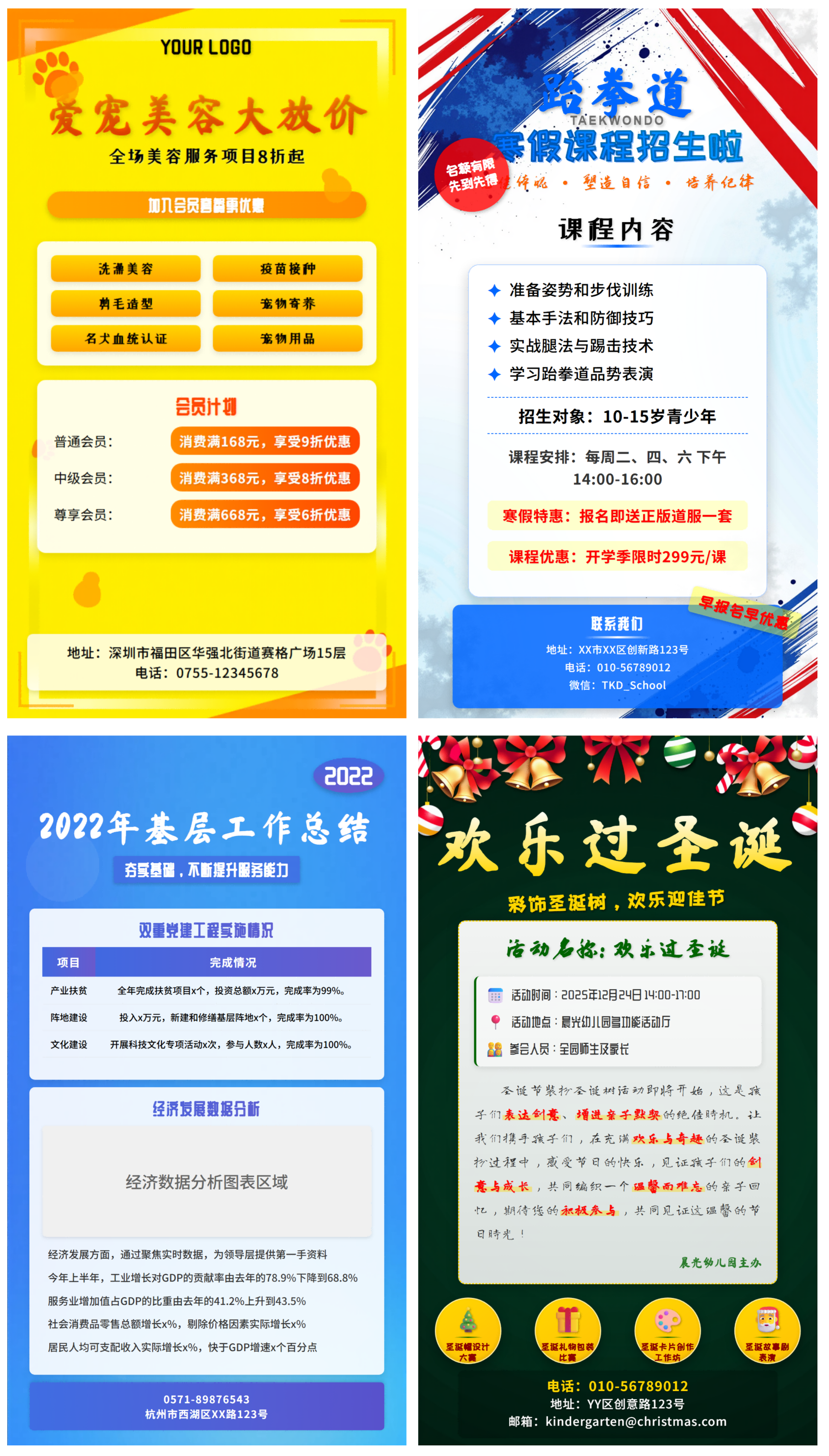}
  \centering
  \caption{ More poster generation results from PosterVerse.}
  \label{fig: show1}
\end{figure*}
\begin{figure*}[t]
  \includegraphics[width=0.75\linewidth]{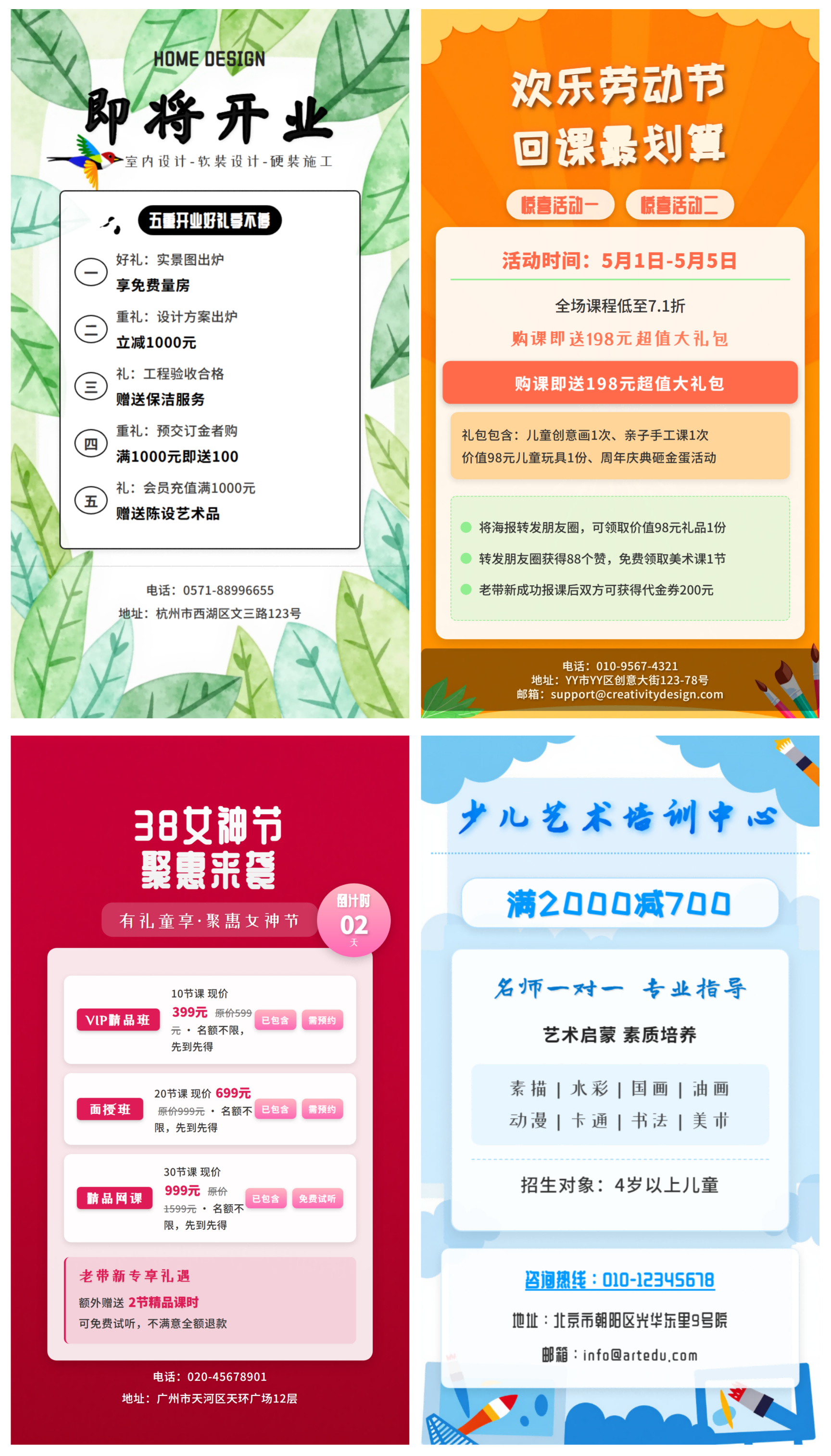}
  \centering
  \caption{ More poster generation results from PosterVerse.}
  \label{fig: show2}
\end{figure*}
\begin{figure*}[t]
  \includegraphics[width=0.75\linewidth]{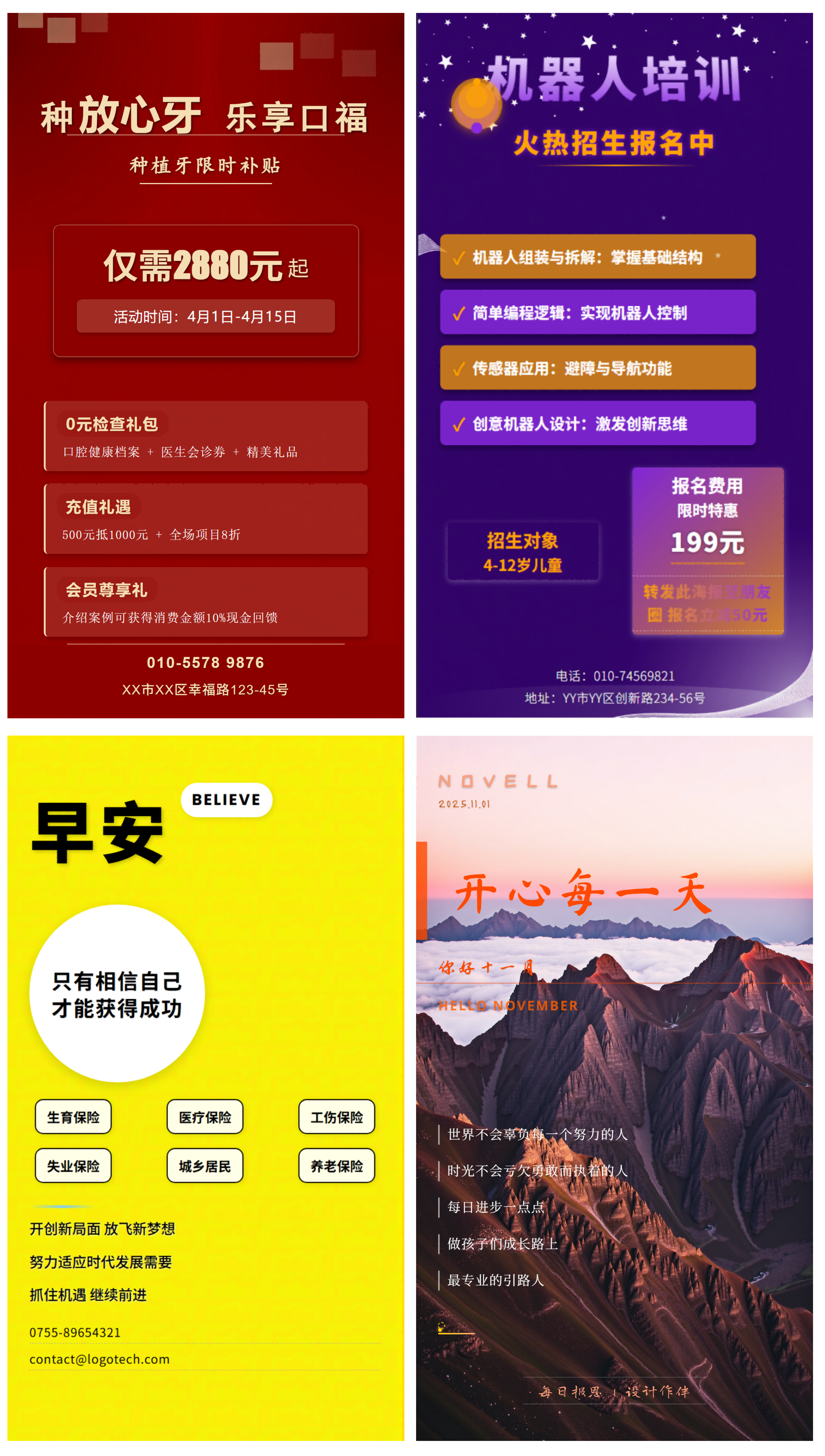}
  \centering
  \caption{ More poster generation results from PosterVerse.}
  \label{fig: show3}
\end{figure*}

\begin{figure*}[t]
  \includegraphics[width=0.75\linewidth]{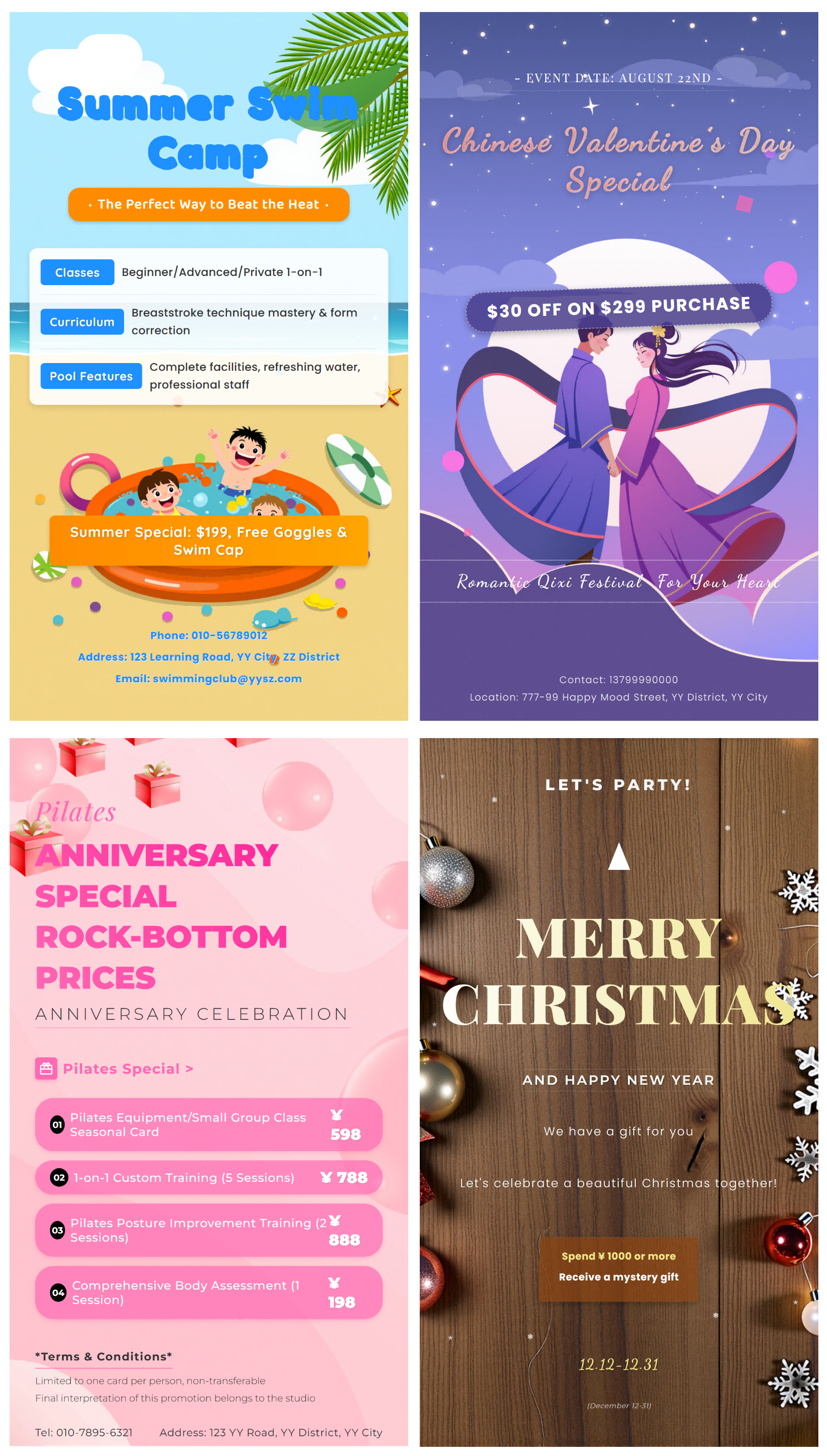}
  \centering
  \caption{Examples of the English posters generated by PosterVerse.}
  \label{fig: english}
\end{figure*}

\begin{figure*}[t]
  \includegraphics[width=0.75\linewidth]{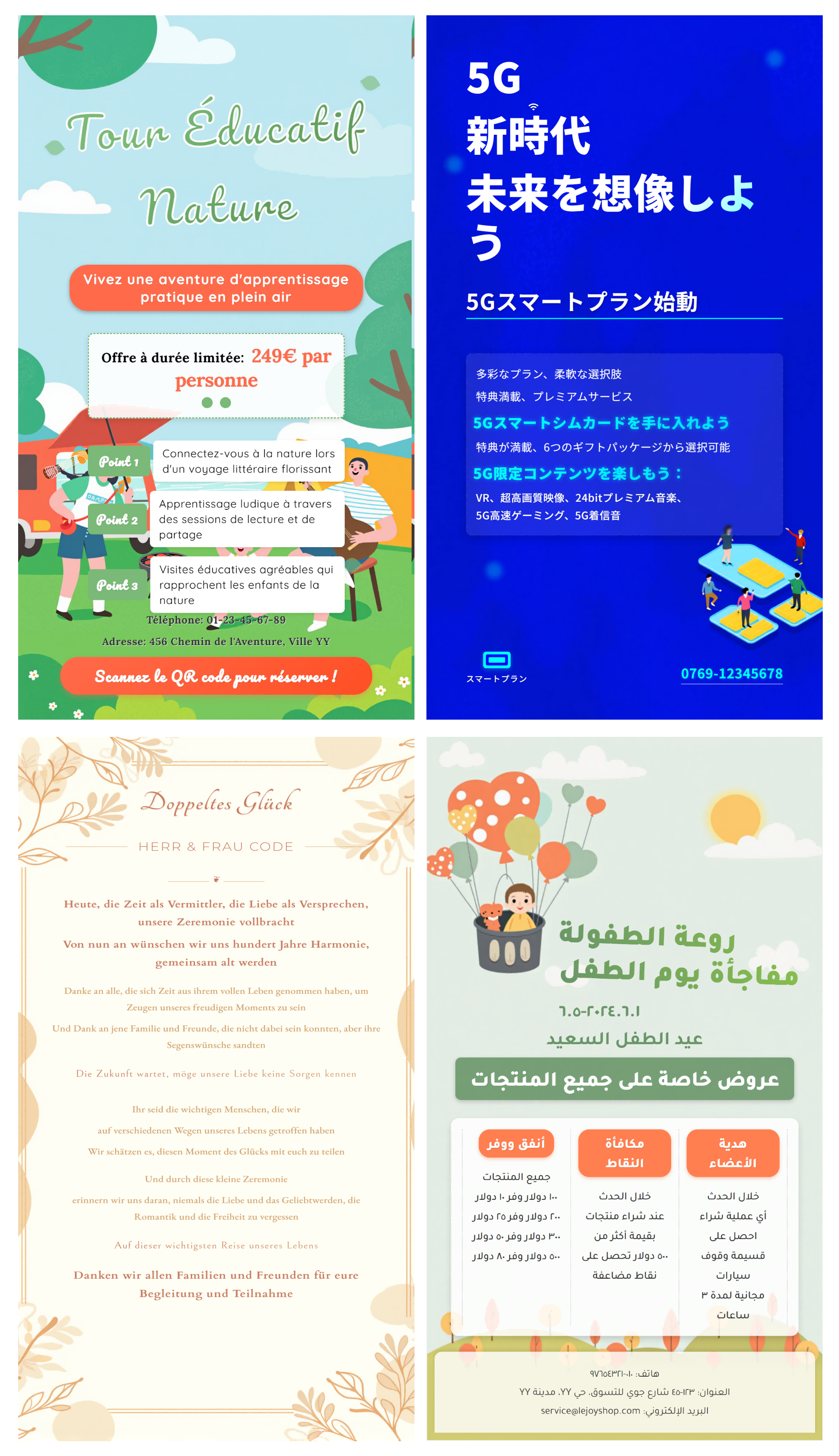}
  \centering
  \caption{Examples of multilingual posters generated by PosterVerse. Top-left: French; Top-right: Japanese; Bottom-left: German; Bottom-right: Arabic.}
  \label{fig: other}
\end{figure*}

\begin{figure*}[t]
  \includegraphics[width=1\linewidth]{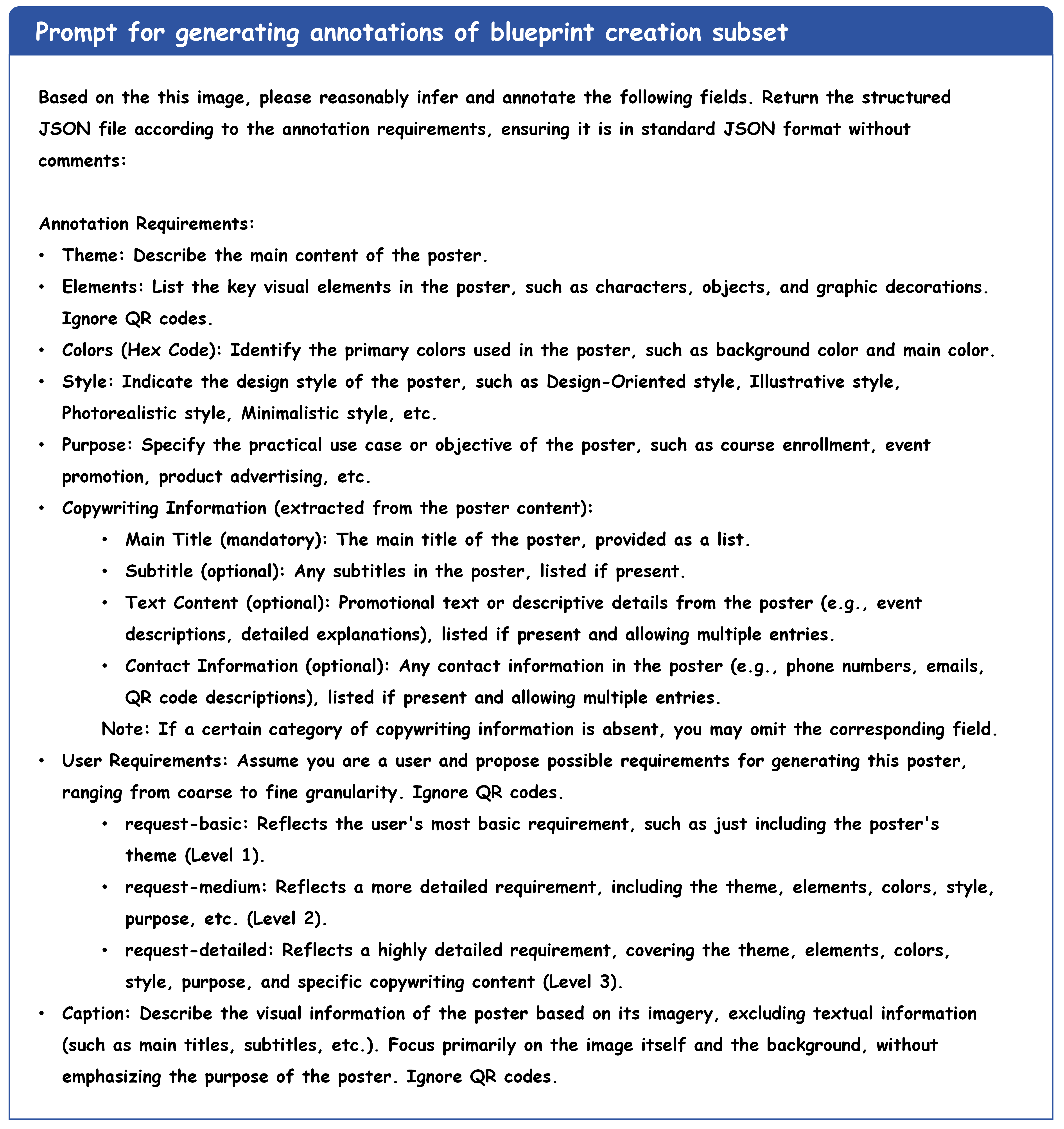}
  \centering
  \caption{The prompt for generating the annotation of blueprint creation subset.}
  \label{fig: prompt1}
\end{figure*}

\begin{figure*}[t]
  \includegraphics[width=1\linewidth]{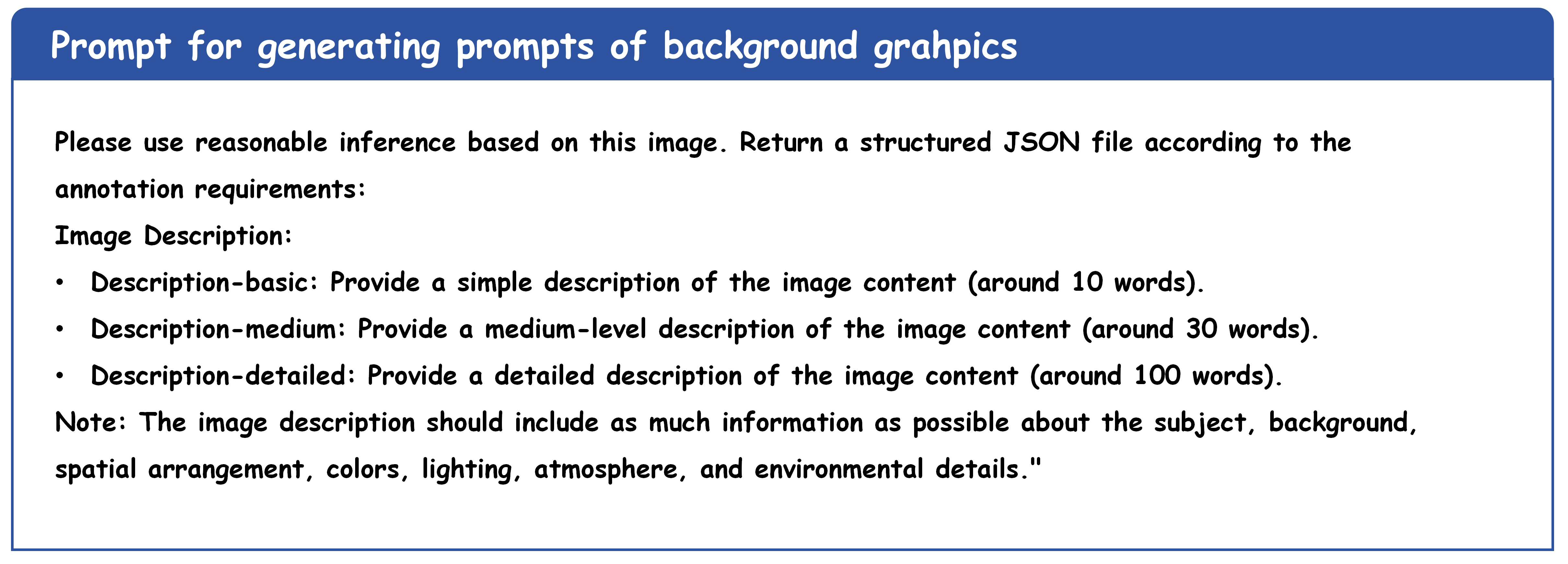}
  \centering
  \caption{The prompt for generating the annotation of the graphic generation subset.}
  \label{fig: prompt2}
\end{figure*}

\begin{figure*}[t]
  \includegraphics[width=1\linewidth]{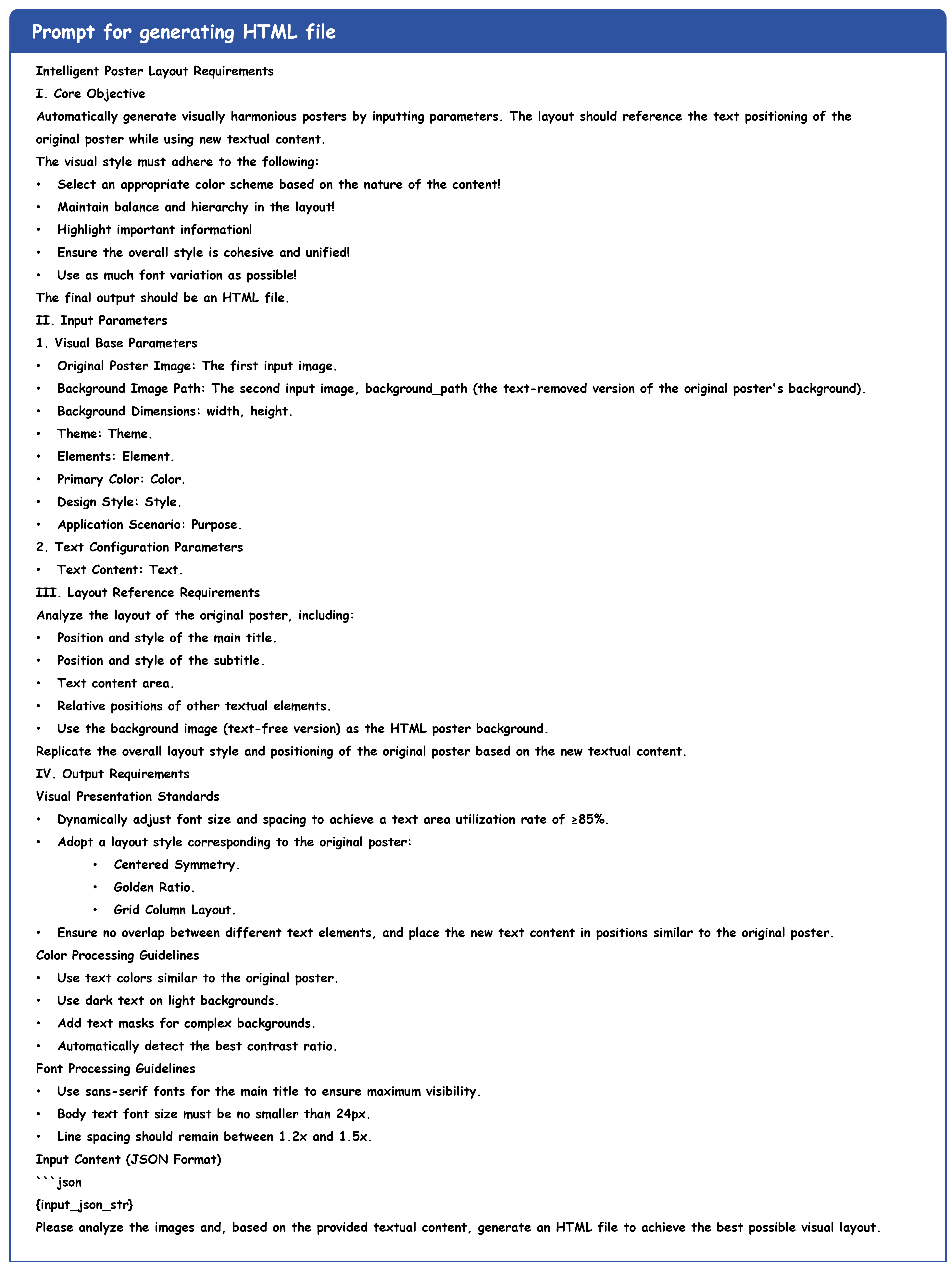}
  \centering
  \caption{The prompt for generating the annotation of the unified layout-text rendering subset.}
  \label{fig: prompt3}
\end{figure*}

\subsection{Data Annotation}
The data annotation process was conducted using Claude 3.7 Sonnet~\cite{claude}. For the blueprint creation subset, the annotation process was guided by the prompt template illustrated in Fig.~\ref{fig: prompt1}, while the graphic generation subset utilized the prompt template depicted in Fig.~\ref{fig: prompt2}. Similarly, the annotation of the unified layout-text rendering subset was based on the prompt template shown in Fig.~\ref{fig: prompt3}. These carefully crafted prompts were specifically designed to address the distinct characteristics and requirements of each subset, ensuring the reliability and precision of the annotated data.

\subsection{Comparison with Existing Poster Datasets}
The comparison between PosterDNA and existing open-source datasets is shown in Tab.~\ref{tab:compareapp}. This comparison highlights the following advantages of PosterDNA. 
First, PosterDNA has the largest dataset size, with 167,000 instances, significantly surpassing other datasets such as the CGL dataset (61,548 instances) and YouTube (11,000 instances), providing a more comprehensive resource for research.
Second, unlike many existing datasets, Second, unlike many existing datasets, PosterDNA offers full workflow coverage data, enabling comprehensive poster analysis and generation tasks. In contrast, datasets such as CGL, PKU PosterLayout, and QB-Poster lack this capability, as they are limited to the generation of poster layouts.
Third, PosterDNA is an HTML-based dataset, which not only promotes modularity and flexibility in design representation but also unifies layout and text rendering. 
This ensures both consistent and precise generation of visual structure and accurate rendering of textual content, a capability absent from existing datasets.
These advantages position PosterDNA as a highly versatile and robust dataset, supporting a wider range of applications and research compared to existing open-source datasets. 
\begin{table}[t]
	\centering
	\begin{tabular}{lcc}
		\toprule
		\textbf{Parameter} & \textbf{Stage 1}\\
		\midrule
		Batch size (per device) & 1 \\
		Gradient accumulation steps & 2 \\
		Learning rate & 1e-5 \\
		Scheduler & Cosine \\
		Epochs & 15 \\
		Sequence length cutoff & 4096 \\
            Float-point precision & bfloat16 \\
		\bottomrule
	\end{tabular}
    \caption{Hyperparameters for blueprint creation training.}
    \label{tab:parameters1}
\end{table}

\begin{table}[t]
	\centering
	\begin{tabular}{lcc}
		\toprule
		\textbf{Parameter} & \textbf{Stage 2}\\
		\midrule
		Batch size (per device) & 1 \\
            Gradient accumulation steps & 4 \\
		  Rank & 64 \\
		Learning rate & 5e-4 \\
		Warmup steps & 10 \\
		Scheduler & Constant \\
		Epochs & 50 \\
            Float-point precision & bfloat16 \\
		\bottomrule
	\end{tabular}
    \caption{Hyperparameters for graphical background generation training.}
    \label{tab:parameters2}
\end{table}

\begin{table*}[t]
\centering
\begin{tabular}{lccccc}
\hline
Dataset          & Venue & \#Instance & \multicolumn{1}{c}{Layout} & \multicolumn{1}{c}{{Full Workflow Coverage}} & \multicolumn{1}{c}{HTML-Base} \\ \hline
CGL dataset~\cite{CGL}      &  arXiv'22     & 61,548     & \ding{51}   & \ding{55}  &  \ding{55}     \\
PKU PosterLayout~\cite{posterlayout} &  CVPR'23     & 10,879     & \ding{51}   & \ding{55}     & \ding{55}      \\
Ad Banner~\cite{layoutdetr}        & ECCV'24      & 8,672      & \ding{51}   &  \ding{55}    &  \ding{55}     \\
YouTube~\cite{youtube}          &  MM'23     & 11,000     & \ding{51}   &  \ding{55}    &  \ding{55}     \\
QB-Poster~\cite{posterllava}        & arXiv'24 & 5,188      & \ding{51}   &  \ding{55}    & \ding{55}      \\
POSTA~\cite{posta}            &  CVPR'25 & 4,500+     & \ding{51}   & \ding{51}      &  \ding{55}     \\ \hline
PosterDNA (Ours) &   -    & 167,000    & \ding{51}   &  \ding{51}    &  \ding{51}     \\ \hline
\end{tabular}
\caption{Comparison with existing open-source datasets.}
\label{tab:compareapp}
\end{table*}

\section{More Experimental Details}
\subsection{Implementation Details}

\subsubsection{Model Implementation}
For blueprint creation, we conducted full-parameter Supervised Fine-Tuning (SFT) training on the Qwen-2.5-14B-Instruct model. The specific hyperparameter configurations are shown in Tab.~\ref{tab:parameters1}. The training was performed on 8 H800 GPUs and took approximately 30 hours to complete. For graphical background generation, we adopted the LoRA approach for training on Flux.1 dev. The specific hyperparameter configurations are detailed in Tab.~\ref{tab:parameters2}. We trained four models on 8 H800 GPUs, with the following time durations: Illustrative Style (approximately 40 hours), Design-Oriented Style (around 58 hours), Minimalistic Style (approximately 96 hours), and Photorealistic Style (roughly 70 hours). For unified text-layout creation, we conducted SFT training on Qwen2.5-VL-7B-Instruct. The specific hyperparameter configurations are detailed in Tab.~\ref{tab:parameters3}. The training was performed on 8 H800 GPUs and took approximately 30 hours.

\begin{table}[t]
	\centering
	\begin{tabular}{lcc}
		\toprule
		\textbf{Parameter} & \textbf{Stage 3}\\
		\midrule
		Batch size (per device) & 2 \\
		Gradient accumulation steps & 8 \\
		Learning rate & 1e-5 \\
		Weight decay & 0.01 \\
		Scheduler & Cosine \\
		Epochs & 50 \\
		Sequence length cutoff & 4096 \\
            Freezing Vision Transformer & False \\
            Float-point precision & bfloat16 \\
		\bottomrule
	\end{tabular}
    \caption{Hyperparameters for unified text-layout creation training.}
    \label{tab:parameters3}
\end{table}

\subsubsection{Evaluation Metric}
Following previous methods~\cite{hiercode, postercraft, posta}, we use the correct rate (CR) and F1 score to evaluate text rendering accuracy. The formulas for CR and F1 are as follows:
\begin{equation}
% AR &= (N_t - D_e - S_e - I_e)/N_t, \\
CR = (N_t - D_e - S_e)/N_t, 
\end{equation}
\begin{equation}
F1 = 2 \cdot \frac{Precision \cdot Recall}{Precision + Recall},
\end{equation}
where $N_t$ is the total number of characters in annotations, while $D_e$, $S_e$, and $I_e$ denote deletion, substitution, and insertion errors, respectively. Precision and Recall are calculated based on the recognition results and ground truth annotations.
Additionally, well-designed layouts typically minimize element overlaps. Following~\cite{posterlayout}, we adopt the overlap metric to quantitatively assess the extent of overlap among elements:
\begin{equation}
Overlavp = \sum_{i=1}^{N} \sum_{j \neq i} \frac{s_i \cap s_j}{s_i},
\end{equation}
where $s_i \cap s_j$ denotes the overlapping area between element $i$ and $j$. $N$ is the total number of elements.

\section{Additional Quantitative Results}
To comprehensively evaluate the performance of each stage in PosterVerse, we employ both Edit Distance and BERT Score as metrics to assess the similarity between the generated result (JSON format) and the ground-truth in the blueprint creation (first) stage, as well as between the generated result (HTML format) and the ground-truth in the unified layout-text rendering (third) stage. As shown in Tab.~\ref{tab:stage1} and Tab.~\ref{tab:stage3}, our model achieves superior results compared to Claude and the Qwen series of models. Additionally, we use FID and CLIP-Image Similarity to evaluate the quality of background images generated in the graphical background generation (second) stage. As shown in Tab.~\ref{tab:stage2}, our model outperforms the current state-of-the-art open-source models, Flux~\cite{flux} and CogView4~\cite{cogview}. These comprehensive evaluation results demonstrate that our model achieves state-of-the-art performance across all stages of the poster generation pipeline, validating the effectiveness of our multi-stage approach.

\begin{table}[t]
\begin{tabular}{lcc}
\hline
\multicolumn{1}{c}{Methods} & Edit Distance ↓ & BERT Score ↑ \\ \hline
Qwen2.5-14B    &   21.90      &  87.75          \\
Claude 3.7 Sonnet                      & 22.32       & 89.50      \\
PosterVerse (Ours)                        & \textbf{12.00}       & \textbf{93.56}     \\ \hline
\end{tabular}
\centering
\caption{Quantitative comparison of blueprint creation quality using Edit Distance and BERT Score.}
\label{tab:stage1}
\end{table}

\begin{table}[t]
\begin{tabular}{lcc}
\hline
\multicolumn{1}{c}{Methods} & FID ↓ & CLIP-IS ↑ \\ \hline
Cogview4    &     60.46     &   83.34          \\
Flux.1 dev        &     40.65     &   86.85         \\
PosterVerse (Ours)        &     \textbf{26.8}     &   \textbf{87.67}  \\ \hline
\end{tabular}
\centering
\caption{Quantitative comparison of graphical background generation quality using FID and CLIP-Image Similarity metrics.}
\label{tab:stage2}
\end{table}

\begin{table}[t]
\begin{tabular}{lcc}
\hline
\multicolumn{1}{c}{Methods} & Edit Distance ↓ & BERT Score ↑ \\ \hline
Qwen2.5-VL-7B         &  68.44     & 91.66      \\
Claude 3.7 Sonnet                      &  67.21     & 90.83      \\
PosterVerse (Ours)                        & \textbf{66.01}       & \textbf{92.41}    \\ \hline
\end{tabular}
\centering
\caption{Evaluation of HTML generation quality in the unified layout-text rendering stage using Edit Distance and BERT Score.}
\label{tab:stage3}
\end{table}

%We also applied our method to other languages, with a visualization example in English shown as illustrated in the Fig.~\ref{}

\section{Additional Qualitative Results}
More poster generation results from PosterVerse are shown in Fig.~\ref{fig: show1}, Fig.~\ref{fig: show2} and Fig.~\ref{fig: show3}. When a poster serves purposes such as notification, recruitment, or sales, it cannot solely focus on aesthetic appeal but must emphasize the effective transmission of textual information. We observe that PosterVerse consistently produces results that meet commercial-grade design standards, validating our framework's capability for professional-quality poster generation. Additionally, PosterVerse demonstrates exceptional performance in textual accuracy and dense text layout, manifested in several aspects. (1) PosterVerse identifies key points within dense text and applies appropriate emphasis. (2) PosterVerse effectively categorizes and organizes dense text into readable and aesthetically pleasing typographic arrangements within the poster. (3) PosterVerse augments the poster with semantic-relevant iconographic elements based on textual content, enriching the visual composition of the poster.

\subsection{Generation Proficiency in Other Languages}
Our method is inherently capable of generating multilingual posters, thus meeting a wider variety of user requirements. This is achieved by only necessitating the translation of user requirements into Chinese, while preserving the original language of the text to be rendered. As illustrated in Fig.~\ref{fig: english}, our model effectively produces English posters, maintaining both high layout quality and textual aesthetics. PosterVerse demonstrates significant advantages in arranging dense English text, achieving aesthetically pleasing typography even without specific training on English language content, thus highlighting the generalization capabilities of PosterVerse. Beyond English, we have also experimented with other languages such as French, Japanese, German, and Arabic. Some examples are shown in Fig.~\ref{fig: other}, demonstrating PosterVerse's capabilities across diverse linguistic contexts.

\end{document}